\documentclass[journal]{IEEEtran}
\usepackage[switch]{lineno}
\usepackage{cite}
\ifCLASSINFOpdf
\usepackage[pdftex]{graphicx}
\else
\fi
\usepackage[cmex10]{amsmath}

\usepackage{makecell}
\usepackage{multirow}
\usepackage{colortbl}
\usepackage{soul}
\definecolor{lightred}{rgb}{0.98,0.91,0.89}
\newcommand*\rot{\rotatebox{90}}

\usepackage[caption=false,font=footnotesize]{subfig}
\usepackage{amssymb}
\usepackage{enumitem}

\usepackage{tikz}
\usepackage{textcomp}

\newcommand\copyrighttext{%
  \footnotesize \textcopyright 2021 IEEE. Personal use of this material is permitted. Permission from IEEE must be obtained for all other uses, in any current or future media, including reprinting/republishing this material for advertising or promotional purposes, creating new collective works, for resale or redistribution to servers or lists, or reuse of any copyrighted component of this work in other works.}
\newcommand\copyrightnotice{%
\begin{tikzpicture}[remember picture,overlay]
\node[anchor=south,yshift=4pt] at (current page.south) {\fbox{\parbox{\dimexpr\textwidth-\fboxsep-\fboxrule\relax}{\copyrighttext}}};
\end{tikzpicture}%
}

\begin{document}
\title{SEnSeI: A Deep Learning Module for Creating Sensor Independent Cloud Masks}

\author{Alistair~Francis,
        John~Mrziglod,
        Panagiotis~Sidiropoulos~\IEEEmembership{Member,~IEEE},
        and~Jan-Peter~Muller
\thanks{Manuscript received May 24, 2021; revised July 22, 2021 and October 29, 2021; accepted November 9, 2021}
\thanks{A. Francis was with Mullard Space Science Laboratory, University College London, and is now with $\Phi$-lab, ESRIN. e-mail: alistair.francis@esa.int}
\thanks{P. Sidiropoulos and J.-P. Muller are with Mullard Space Science Laboratory, University College London}% <-this % stops a space
\thanks{J. Mrziglod was previously with the World Food Programme, Rome}% <-this % stops a space
}


\markboth{}%
{Shell \MakeLowercase{\textit{et al.}}: Bare Demo of IEEEtran.cls for Journals}



\IEEEoverridecommandlockouts
%\IEEEpubid{\makebox[\columnwidth]{\copyright2021 IEEE. Personal use of this material is permitted. Permission from IEEE must be obtained for all other uses, in any current or future media, including reprinting/republishing this material for advertising or promotional purposes, creating new collective works, for resale or redistribution to servers or lists, or reuse of any copyrighted component of this work in other works.\hfill} \hspace{\columnsep}\makebox[\columnwidth]{ }}

\maketitle
\copyrightnotice

%\IEEEpubidadjcol


\begin{abstract}
We introduce a novel neural network architecture---Spectral ENcoder for SEnsor Independence (SEnSeI)---by which several multispectral instruments, each with different combinations of spectral bands, can be used to train a generalised deep learning model. We focus on the problem of cloud masking, using several pre-existing datasets, and a new, freely available dataset for Sentinel-2. Our model is shown to achieve state-of-the-art performance on the satellites it was trained on (Sentinel-2 and Landsat 8), and is able to extrapolate to sensors it has not seen during training such as Landsat 7, Per\'uSat-1, and Sentinel-3 SLSTR. Model performance is shown to improve when multiple satellites are used in training, approaching or surpassing the performance of specialised, single-sensor models. This work is motivated by the fact that the remote sensing community has access to data taken with a hugely variety of sensors. This has inevitably led to labelling efforts being undertaken separately for different sensors, which limits the performance of deep learning models, given their need for huge training sets to perform optimally. Sensor independence can enable deep learning models to utilise multiple datasets for training simultaneously, boosting performance and making them much more widely applicable. This may lead to deep learning approaches being used more frequently for on-board applications and in ground segment data processing, which generally require models to be ready at launch or soon afterwards.
\end{abstract}


% \fi
\IEEEpeerreviewmaketitle

\section{Introduction}
\label{sec:intro}

\IEEEPARstart{T}{here} is an ever-increasing quantity of remote sensing data---supplied by hundreds of satellites---being captured and downlinked every day. To preprocess and analyse this satellite data, automated techniques are essential. Deep learning has been of particular interest to remote sensing practitioners recently~\cite{Zhu2017deep, Ma2019deep}, given its proven abilities in a wide range of computer vision problems. Cloud masking is one such instance, which we will focus on specifically in this paper. However, many others exist, e.g. crop-type classification~\cite{Orynbaikyzy2019crop}, sea ice detection~\cite{Zakhvatkina2012seaice}, forest fire detection~\cite{Tanase2020burned}, urban structure analysis~\cite{Wurm2009urban}, and sea vessel tracking~\cite{Kanjir2018vessel}.

When these tasks are approached through the lens of supervised learning, labelled datasets are needed for the dual purpose of training and testing predictive models (often deep learning models, which are our focus here, but not exclusively). The labour required to annotate datasets is substantial in all fields of computer vision~\cite{Roh2019survey}, but is perhaps particularly relevant in remote sensing, because each dataset is specific to not only a given problem, but also a given sensor. In practical terms, this has led to a somewhat fractured set of data collection projects, each producing relatively modest datasets that target a specific satellite. These cannot easily be combined to add cumulative value to a predictive model, although exceptions to this do exist, for example the many high-resolution satellites that use similar sensor designs with RGB channels only, often using Bayer color filter patterns, which lend themselves to direct interoperability. 

These atomised data collection efforts mean that when a researcher approaches a problem, one of two things often happen. Many decide to work on the largest, highest quality dataset available for a given problem, which can lead to an over-focus of studies on a few satellites and datasets. Alternatively, if a researcher deems it necessary, they will resort to creating their own purpose-built dataset for their sensor of choice. This is by no means an entirely negative consequence, as those datasets can add value to others' work if made publicly available. However, it does mean that the prior efforts of other researchers---who may have created labelled datasets and designed models for the same task but on different sensors---are not adding value to the project. In this way, the quantity of labelled datasets used in model training seems unlikely to dramatically increase, even as the total quantity of existing labelled data does continue to grow.

Given that deep learning's potential is most easily realised with huge quantities of training data, we argue that researchers should continue to examine ways to build interoperability into their work, in order to make the quantity of usable labelled data increase in proportion to the total labour expended on annotation, and lead to larger and larger amounts of publicly available training data. 

In pursuit of this interoperability, we define \textit{sensor independence} as the ability of a model to ingest data from different multispectral sensors, without retraining, and without the need to preprocess the data to be in the same format. A sensor independent model is one which can take data without a fixed number of spectral bands. This is related to---but distinct from---the concept of multimodality, which concerns the \textit{combination} of data from multiple sensors into a single input (e.g.~\cite{Gomez2015multimodal,Hong2020more}), whereas a sensor independent model tries to perform a task with data from one sensor, regardless of which sensor it is.

Sensor independence is a skill that humans largely take for granted given the cognitive ease with which we can assimilate and compare visual inputs of very different kinds, e.g. our ability to simultaneously understand contour plots and heatmaps. Our ability to do this hinges on our understanding of how these different styles can be translated to express the same underlying concepts. Similarly, for a deep learning model to emulate this capability, a translation from the individual satellite representations into a shared, general representation must be made.

Besides an expansion in the available training data for a model, sensor independence also makes deep learning approaches more practicable, and immediately usable on a wider variety of sensors. Algorithms used for data processing by satellite operators are usually designed before launch, so that they can be applied immediately when the satellite begins to downlink data. Machine learning and deep learning techniques---despite commonly exhibiting better results than other methods---are rarely used in this setting. This is often because without access to training data prior to launch, it is not easy to develop a model which relies on supervised learning. A similar situation exists for on-board deployment of machine learning. Both cases would be made more practical if models could be developed prior to launch, using pre-existing datasets from other satellites, which sensor independence would allow.

This paper proposes a versatile framework, Spectral ENcoder for SEnsor Independence (SEnSeI), which enables deep learning architectures to become sensor independent.  SEnSeI is a pre-processing module, which acts as the translator between the ``languages'' of individual sensors (their individual channels' reflectances and wavelengths) and a shared feature space that can then be inputted into an existing deep learning model (or indeed, any other machine learning model). SEnSeI's design does not depend on the model it is used with, allowing it to be bolted onto any pre-existing machine learning or deep learning architecture, which can then be trained to be sensor independent. SEnSeI is designed as a permutation invariant neural network, the theoretical background and practical development of which is set out in Section \ref{subsec:perminv}. 

Ideally, the addition of SEnSeI to a model will have no deleterious effects on its performance, whilst also making it sensor independent. In order to distil down the implications of SEnSeI, we tie all our experimental results back to the following hypotheses:

\begin{enumerate}[align=left]
	\item [\textit{\textbf{Hyp. (A):}}] {If SEnSeI is added to a deep learning model, there is no reduction in performance when trained and tested on data from a single sensor.}
	\item [\textit{\textbf{Hyp. (B):}}] {If SEnSeI is added to a deep learning model, the model can be trained on multiple sensors without a drop in performance relative to when it is trained on an individual sensor.}
	\item [\textit{\textbf{Hyp. (C):}}] {SEnSeI can enable a model to perform adequately on a previously unseen sensor, given training data that provide close equivalents to the unseen sensor's spectral bands.}
\end{enumerate}

In order to measure SEnSeI's performance and ability to generalise to different sensors, we have chosen cloud masking as a case-study. Cloud masking is a central problem in Earth Observation, as rejecting cloudy areas is a vital pre-processing stage in many other applications. It is well studied, and has large datasets from several sensors, but these datasets have not yet allowed researchers to employ deep learning to cloud masking on other sensors which don't currently have large manually labelled datasets. This is a key limitation in current approaches, because whilst high performance can be demonstrated on individual sensors, the techniques are usually not applicable to other multispectral instruments.

Both a single-pixel neural network, and a large convolutional model (DeepLabv3+~\cite{Chen2018deeplab}) are combined with SEnSeI, using various training set configurations. These two models show how SEnSeI interacts with as wide a range of models as possible, from the very simple to the very complex. Our key findings in this paper include:

\begin{itemize}
	\item{A novel model architecture, SEnSeI, is proposed. SEnSeI is a permutation invariant neural network, with unique features specifically designed for remote sensing data.}
	\item{DeepLabv3+ is shown to work extremely well as a cloud masking model, exhibiting higher performance than other existing techniques for several multispectral satellites.}
	\item{By combining SEnSeI with DeepLabv3+, training datasets can be combined to make models which work across many instruments, with a range of resolutions and spectral band combinations.}
	\item{Whilst in some instances, sensor independence leads to higher numerical performance (e.g. in our tests on Landsat 8 data), the primary benefit of adopting sensor independence with SEnSeI is the increase in a model's usability. With SEnSeI, it is now possible to create reliable cloud masks for a huge family of multispectral instruments, even those which do not currently have labelled datasets.}
\end{itemize}

After a section exploring previous studies pertinent to our work (Section~\ref{sec:rel-work}), our experimental setup, including model design, datasets and their preprocessing, and model training is detailed in Section \ref{sec:method}. Then, quantitative test results on four satellites' data, Sentinel-2, Landsat 8, Landsat 7 and Per\'uSat-1 are given in \ref{sec:results}, along with visual results on Sentinel-3 SLSTR. In Sections \ref{sec:discuss} and \ref{sec:conc} we discuss how these results fit with the aforementioned hypotheses, the possible implications this work has on deep learning in remote sensing, some of the future avenues of research this entails, and its limitations. The code associated with this work is publicly available~\cite{Francis2021SEnSeI}.

\section{Related Work}\label{sec:rel-work}

\subsection{Satellite Interoperability}\label{subsec:interoperability}

Here we consider the similarities and differences between satellite sensors, and prior studies which successfully use combinations of satellites with predictive models. Comparison of data between Sentinel-2 and Landsat 8 has shown good correlation~\cite{Helder2018observations}. By comparing coincident observations between Sentinel-2 and Landsat 8, their top of atmosphere reflectances were shown to be within 1\% (Blue and Coastal Aerosol bands showed slightly more deviation at 1.5\%). A similar analysis over desert sites also found results that suggested good agreement between the instruments~\cite{Barsi2018sentinel}.

A prominent example of data calibration across satellites is NASA's harmonization project~\cite{Claverie2018harmonized}, calibrating Sentinel-2 with Landsat 8. This has allowed for time series analyses with more frequent observations, as more cloud-free images can be found when using multiple sensors. Applications of this harmonized data include detection of irrigated land using several classifiers~\cite{Falanga2020harmonized}, high-resolution land phenology~\cite{Bolton2020continental}, and winter wheat yield estimation~\cite{Skakun2018winter}. Whilst interoperability of contemporaneous missions can lead to more frequent observations, calibration of satellites past and present can lead to longer baseline studies, as~\cite{Banskota2014forest} covers in the context of forest monitoring.

These approaches show that interoperability can be a highly successful strategy for expanding available Earth Observation data, by lowering revisit times, or expanding the period of study. A general trend that is seen in current applications of interoperability is shared feature selection; derived values, e.g. the Normalized Difference Vegetation Index (NDVI)~\cite{Rouse1974NDVI} are calculated for the different sensors. This effectively translates the instruments into a shared representation which a predictive model can use. We propose instead to make this translation into a general, non-linear process, with SEnSeI. This comes with the hypothetical advantage of the predictive model's utilisation of \textit{all} spectral information in each available satellite, not only those bands used to create derived values such as NDVI.

\subsection{Cloud Masking}\label{subsec:cloudmasking}

A wide variety of modelling approaches have been studied in the detection of clouds~\cite{Mahajan2019cloud}. Some models have used multitemporal information to predict cloud cover (e.g.~\cite{Champion2012automatic,Zhang2020thick, Zhang2021combined}). By leveraging the differences between images in a time series, one can ascertain which images contain cloud. However, we focus primarily on those which take single images as their input, as that is the approach we prefer.
 
Many algorithms are based on reflectance or radiance thresholds, which are specified by experts and are usually fixed. For example, Landsat 7's ACCA~\cite{Irish2006characterization} estimates the cloud cover over a scene by passing the band values through a rules-based system, which predict whether each pixel is a cloud. \cite{Scaramuzza2011CCA} builds and improves upon this design with AT-ACCA. Similarly,~\cite{Bley2013threshold} use a thresholding method on the high-resolution visible channel of Meteosat SEVIRI. Sen2Cor, which serves to correct for atmospheric effects---including clouds---in Sentinel-2 Level 2 products also uses a rules-based approach\cite{Louis2016sentinel}. Fmask~\cite{Zhu2012fmask, Zhu2015improvement} is an algorithm originally designed for Landsat imagery, which uses situational thresholding based on whether the cloud is thought to be above water or land, and can assign confidence values to its predictions, allowing users to balance their needs for sensitivity and specificity of detections. Unfortunately, in studies comparing the relative performance of different methods, thresholding based approaches tend to fall short of techniques which embrace some kind of optimisation through labelled training data~\cite{Mahajan2019cloud, Foga2017Biome}.

Convolutional Neural Networks (CNNs) have come to dominate research in many Computer Vision tasks, largely triggered by the publication of AlexNet~\cite{Krizhevsky2012imagenet, Alom2018history}. For image segmentation, many architectures have been proposed. For example, U-Net~\cite{Ronneberger2015unet} was developed for biomedical image segmentation, but variations of it have also been used with success in cloud masking~\cite{Francis2019cloudfcn, Wieland2019multi, Jeppesen2019rsnet, Zhang2020lightweight}. More bespoke architectures have also been developed, e.g.~\cite{Mohajerani2019cloud, Mohajerani2020cloud, Khoshboresh2020deep, Molinier2018deepcloud, Li2018cloud}. Given that our work concerns sensor independence, and our contribution, SEnSeI, could theoretically be used with any and all machine learning architectures, we forgo a discussion of their relative benefits, (though such a discussion is available in~\cite{Mahajan2019cloud}). Instead, we now turn to look at studies which have focused on using models across different sensors, as we have done.

Many studies have applied deep learning models to cloud masking across multiple sensors. Here, we define some broad categories of multi-satellite model application that have been previously studied, and differentiate them from our proposed framework of sensor independence. First, we characterise \textit{passive interoperability} as the ability to use the same model architecture across different sensors, but with training datasets for each being needed. This is the most common form of interoperability within deep learning-based cloud masking.

Next, we define \textit{transfer learning} (in the context of cloud masking) as the opportunistic use of similarities between sensors to train models on specific, shared spectral band combinations. This approach still provides a model which is not generalised, in that it cannot then be used on any arbitrary set of spectral bands. This has two drawbacks, (i) the number of sensors a model can work with is limited to only those with all of the selected bands, and (ii) the information from other spectral bands not included in the specific combination of selected bands is discarded. \cite{Shendryk2019deep} use PlanetScope data, and Sentinel-2 data restricted to the RGB and NIR bands, to simulate the same spectral bands as PlanetScope, allowing for a model which is interoperable with 4-band RGB+NIR satellite data. Similarly, \cite{Li2019deep} shows a single model working across RGB images taken from 8 different sensors and a range of resolutions, from 4--50 m. \cite{Wieland2019multi} notes that not only are RGB and NIR bands shared, but also that the two SWIR bands across Sentinel-2, and Landsat 5, 7, and 8 are similar, using this to train a 6-band model shared across the four sensors.

In contrast to passive interoperability and transfer learning, we propose the term \textit{sensor independence} to describe the property of a model which can be used across different sensors without selecting a specific spectral band combination, and without retraining. The bands that a sensor independent model can use are still limited to those available to the model during training, but any combination of those bands can be taken as input. This affords us greater flexibility in sensor choice, more training data, and a more comprehensive leveraging of the available information from the spectral bands those sensors' data include, by not throwing away those bands which are not shared. We now introduce the machine learning concepts which make this possible.

\subsection{Permutation Invariant Neural Networks}\label{subsec:perminv}

A distinct formulation of neural network architectures has arisen, in recent years, for tasks which do not have a fixed number of input variables. Permutation Invariant Neural Networks (PINNs) are a broad family of model designs, which have the property that the output of the network does not change given any re-ordering (permutation) of the set of inputs. The PINN is therefore symmetric with respect to the order of the inputs. This section first introduces the theoretical basis for permutation invariance, and then gives some examples pertinent to our work.

Permutation invariance can be achieved through any symmetric pooling operation (e.g. summation, maximisation, or mean averaging) over the outputs of some function acting on a set of inputs. Broadly, a PINN can be seen as applying a transformation (which in the case of a PINN is a neural network), $\mathcal{F}$, across the inputs at each index $i$ of a set of $N$ inputs $X$, then pooling those $N$ transformed inputs with an operation $\mathcal{P}$.

\begin{equation} \label{eqn:pinn}
y=\mathcal{P}(\{ \mathcal{F}(X,\; i)\; \;  \forall \; \; i  \; \in \; 1, 2, ... \: , N \} )
\end{equation}

This formula has the useful property that the size of the vector \textit{y} is independent of the number of input elements, \textit{N}, allowing for any arbitrary group of inputs to be represented in the same vector space, which a standard neural network can then use.
 
In simple cases, $\mathcal{F}$ acts upon each element $x_i$ individually. However, more complex operations can be performed which rely on other members of the set of inputs $X$, such that elements other than $x_i$ affect the $i^{th}$ output. Therefore, in the general case, if the output $y$ is  always to be symmetric with respect to permutations on the members of $X$, then $\mathcal{F}$ must be \textit{equivariant} with respect to any index swapping transformation, $\mathcal{T}$, on a pair on input indices \textit{i} and \textit{j}. Equivariance, then, means that any reordering of inputs results in exactly the same reordering of outputs:

\begin{equation} \label{eqn:symm}
\mathcal{F}(\mathcal{T}_{ij}(X),\; i) = \mathcal{F}(X,\; j)
\end{equation}

Several studies have proposed PINN models which satisfy this definition. PointNet~\cite{Qi2016point} and PointNet++~\cite{Qi2017point} are prominent examples, designed for the classification or segmentation of point cloud data. They process each 3-dimensional point \textit{(x, y, z)} into a 1024-dimensional feature vector. These are then combined across all $N$ points using a max-pooling operation, to produce a single 1024-dimensional vector which holds information on all points in the point cloud simultaneously.

Permutational layers---which combine each pair of $N$ inputs, and then pool them so as to return $N$ outputs---are permutationally equivariant (satisfying (\ref{eqn:symm})) and have been shown to work well in modelling dynamics~\cite{Guttenberg2016permutation}. Each permutational layer consists of a neural network and allows for information about the interactions between inputs to be considered by the network, by looking at each pair. This allowed the model to successfully predict the velocities and positions of idealised discs moving and colliding in 2 dimensions. 

Unfortunately, pairwise operations have $\mathcal{O}(n^2)$ complexity, and so their application may not be practical in tasks with a large number of input points. To alleviate this, PointNet++~\cite{Qi2017point} limits pairwise processing to only those points within the vicinity of the target point, limiting the computational complexity whilst retaining the fusion of information from nearby points. In our work, this is not a concern, given that the maximum number of spectral bands in multispectral imagery is comparatively low. However, future work concerning sensor independence with hyperspectral data---which contains many more bands---may benefit from an approach which intelligently reduces the number of band pairings considered.

\section{Methods}\label{sec:method}

\subsection{SEnSeI's Design}

SEnSeI is an example of a PINN. By this, we simply mean that the output of SEnSeI is independent of the order of the inputs. In practical terms, this invariance is achieved through a simple averaging over the outputted features corresponding to each input. Imagine a neural network ingesting $N$ input vectors, and thus outputting $N$ feature vectors, which are then pooled to make a single feature vector. This averaged feature space is the shared representation that SEnSeI passes to a standard deep learning model. Importantly, this can only be done because of the fixed size of the representation outputted by SEnSeI; an output which changed shape based on the inputs would not be compatible with standard deep learning models. 

The structure of satellite data is the primary driver for the design of SEnSeI. In essence, satellite data can be considered as a set of channels, each with an extended spatial map of reflectances, and a spectral profile. This set of bands is our set of input points to SEnSeI, and its output is a matrix of the same spatial dimensions as the original bands, but with a fixed number of channels. The precise nature of this representation is, of course, non-linear and determined through stochastic training of the network, meaning the output channels do not necessarily have a well-defined physical analogue, or strong associations with any one input channel.

Essentially then, we want SEnSeI to be able to take each channel, and represent them all in a way that preserves the information about their reflectance and their wavelength, and also ensures that when pooled, that information is not degraded or confused with other channels. If trained on a fixed set of input channels, it is easy to see how this architecture could overfit, given that the spectral profiles it fits to would always be identical. To avoid this, random subsets of the available channels are used to increase variation, as described further in Section \ref{subsec:sens-training}.

\begin{figure*}
\centering
\includegraphics[width=0.95\linewidth , trim=3cm 0.32cm 2.3cm 1cm , clip]{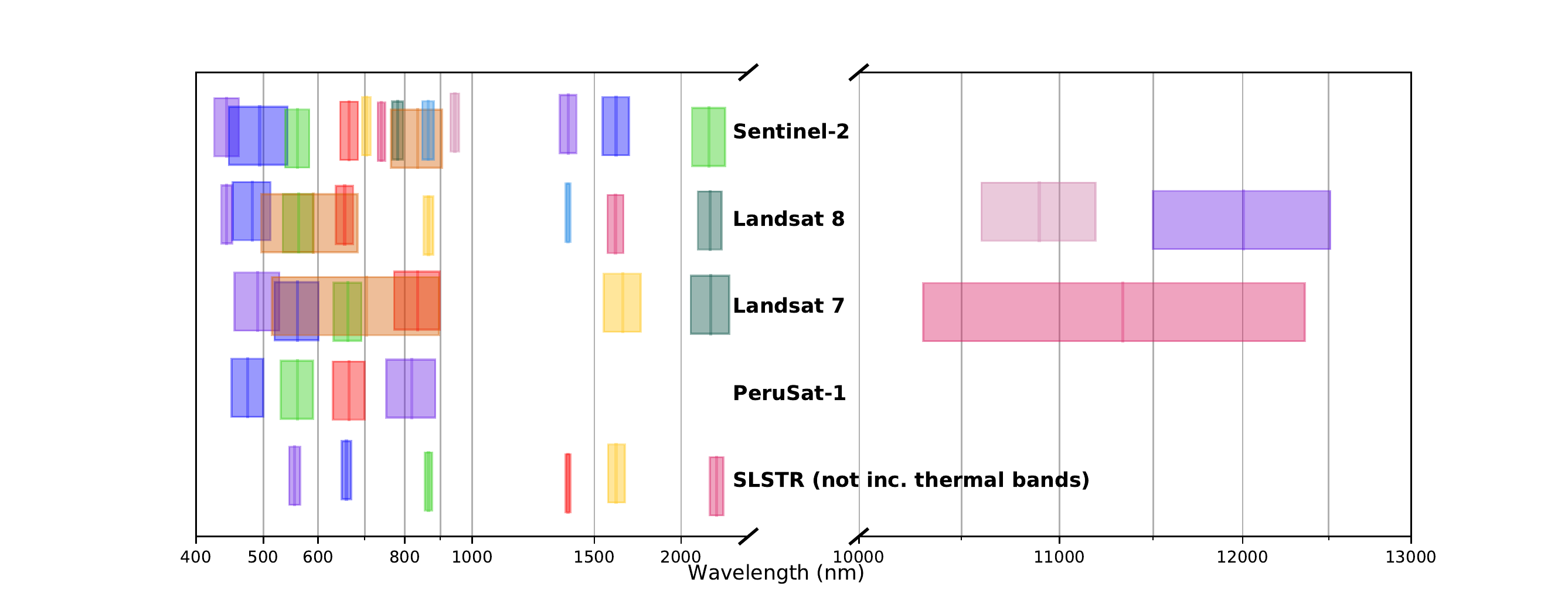}
\caption{The band spectra of the instruments used in this paper. This shows the diversity of sensors used, and underlines the need for sensor independence. The values here represent the FWHM of the bands, but should not be taken as exact. The descriptor vector corresponding to each band comprises its minimum, central, and maximum wavelengths. SLSTR's thermal bands are not shown as we do not use them in our experiments. Colours and small vertical offsets are used arbitrarily, to help make neighbouring and overlapping bands more distinct.}\label{fig:spectra}
\end{figure*}

We parameterise the spectral profile of each band as a `\textit{descriptor vector}'. This consists of a vector with three values: The lower, central and upper wavelengths of the spectral band in question. The wavelength values are normalised, with further details of the procedure found in Section \ref{subsec:preproc}. Fig.~\ref{fig:spectra} visualises the spectral bands of the sensors used in this experiment.

\begin{figure}
\centering
\includegraphics[trim=1.5cm 2.3cm 9.5cm 3cm, clip, width=0.98\linewidth]{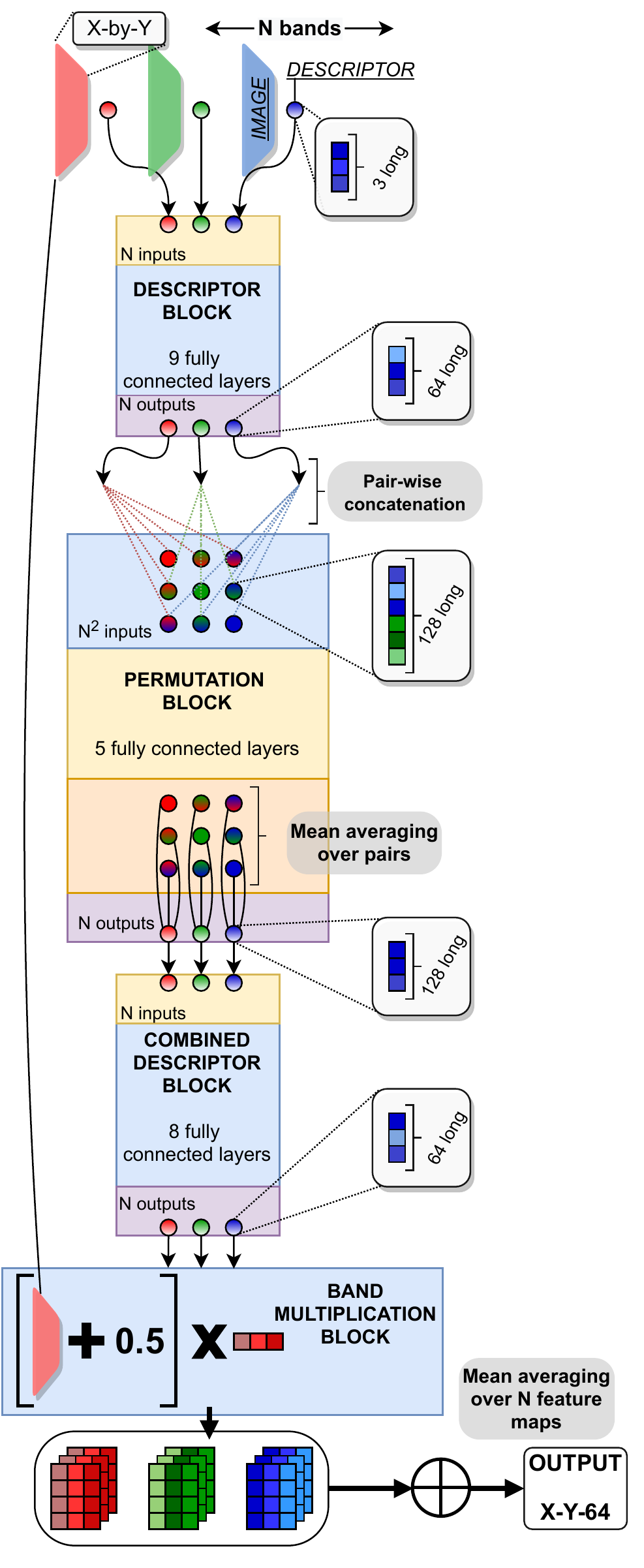}
\caption{Schematic of SEnSeI. Inputs are $N$ images with $N$ corresponding descriptor vectors (circles), with only 3 (red, green and blue) shown here for simplicity. Band multiplication creates feature maps which have the same extent as the input image, wherein each pixel is represented by a 64-dimension vector, scaled by the band's reflectance in that position, plus an offset of 0.5 such that low reflectances are still represented. The entire operation is permutation invariant, given that the final mean averaging over the feature maps does not depend on the inputs' order.}\label{fig:sensei}
\end{figure}

The architecture of SEnSeI can be split into 4 sub-modules, detailed in Fig. \ref{fig:sensei}. First, the `\textit{DESCRIPTOR BLOCK}' comprises a neural network with several fully connected layers, with ReLU and Group Normalization layers between them. In our case, 9 layers were used, with a final output vector of 64 features. At this point, there are $N$ vectors being outputted from the `\textit{DESCRIPTOR BLOCK}', each corresponding to one of the $N$ bands at input. No information from the reflectance values, nor other bands' descriptor vectors, are used in these calculations. Simply, each of the $N$ descriptor vectors (of length 3) enter a 9-layer neural network, which transforms them each into vectors of length 64.

Next, the `\textit{PERMUTATION BLOCK}' takes each possible pair of the $N$ outputs from the previous module, and concatenates them together into $N^2$ 128-dimensional feature vectors, where each band is paired with all $N$ bands, including itself. The 128 dimensions simply come from the two 64-dimensional vectors concatenated together. These are then fed through 5 fully connected layers, and outputted as 128-dimensional vectors. To reduce the number of vectors from $N^2$ back down to $N$, a pooling operation must be performed. By averaging across the $N$ pairs associated with each of the bands, we then reduce the number of feature vectors back down to $N$. 

Third, a `\textit{COMBINED DESCRIPTOR BLOCK}' does essentially the same as the `\textit{DESCRIPTOR BLOCK}' module, but this time acts on the outputs of the previous module, again producing a 64-dimensional feature vector for each of the $N$ inputs. This block is the final set of neural network layers in SEnSeI.

At this stage, SEnSeI has not used any of the reflectance band information. To do so within a neural network would require us to compute the outputs of each layer across each pixel of the input image, as the pixels do not all have the same reflectances. This could become very computationally intensive for large images. To get around this, we instead introduce the reflectance information with the `\textit{BAND MULTIPLICATION BLOCK}'. This module takes the 64-dimensional output feature vectors of the `\textit{COMBINED DESCRIPTOR BLOCK}' and combines it with the X-by-Y matrices of reflectance values for each band. At each position in the original reflectance map, the 64-dimensional feature vector is multiplied by the corresponding reflectance value, with an added offset of 0.5 (so that information from low reflectance values are still included in the output). This produces an N-by-X-by-Y-by-64 feature map. This multiplication is computationally cheap and so can be carried out easily over each pixel.

The entire operation, up to this point, can be thought of as our transformation $\mathcal{F}$ from (\ref{eqn:pinn}). If we were to swap any two inputs, say, Red and Green, in Fig. \ref{fig:sensei}, then we can see that the order of the outputs would be changed in the same way, hence satisfying (\ref{eqn:symm}). Finally, we apply our pooling operation, $\mathcal{P}$, to our set of feature maps: A mean averaging over the $N$ feature maps, producing an X-by-Y-by-64 feature space that can be directly inputted into any other model, in our case DeepLabv3+. Whilst SEnSeI does not fuse information between pixels, the spatial features one finds in the input image are also present in the outputted feature map, because each pixel's feature vector is affected by its input reflectances. This means that models attached to SEnSeI still have access to that spatial information.

Although this approach is applied specifically to multispectral data, SEnSeI could easily be adapted to work with a range of different data types. Essentially, any data structure for which there is a set of images, each with a corresponding group of descriptors (in this case wavelengths) that can be parameterised.

Whilst SEnSeI has many neural network layers, computation is incredibly cheap, because the neural networks act only on the descriptor vectors, rather than the much larger image data. The final band multiplication operation is the only one which scales with the spatial dimensions of the image data. During supervised training we saw only a ~5\% slow down per iteration compared to the standard DeepLabv3+ model. When using the models to make predictions after training, SEnSeI did lower the speed by a small amount, but only when many bands were used (see Appendix~\ref{appendix:computation}).

\subsection{SEnSeI's Pre-training}
\label{subsec:sens-training}

The weights for SenSeI were pre-trained before the supervised training of cloud masks, as part of an autoencoder architecture.  There were two primary aims in this exercise: (i) optimise SEnSeI's design, by testing many different architectures, and (ii) pre-train the weights of SEnSeI for use later in the supervised training of SEnSeI with DeepLabv3+.

\begin{figure}
\centering
\includegraphics[width=0.98\linewidth]{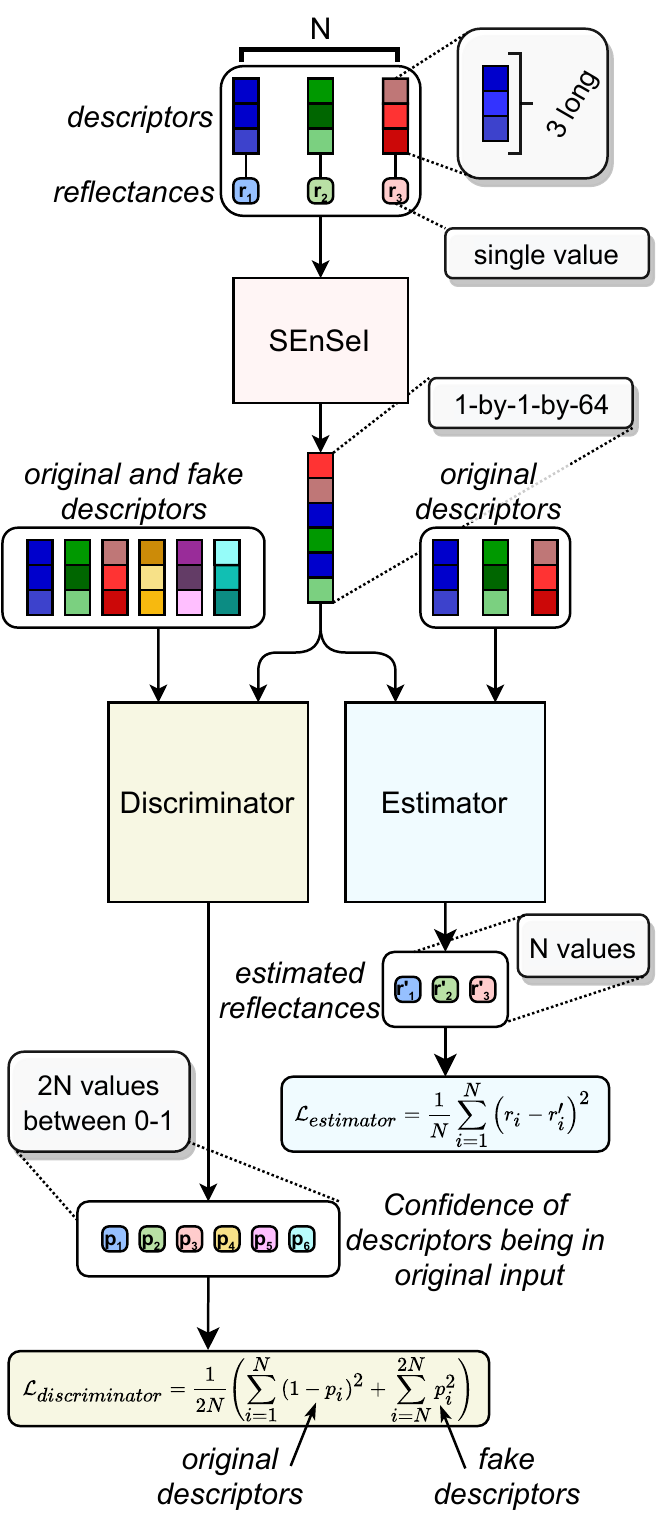}
\caption{Flowchart for unsupervised autoencoder training of SEnSeI. A set of $N$ inputs comprising pairs of random descriptor vectors and reflectance values are processed by SEnSeI. Then, two decoder branches, the discriminator and estimator, attempt to recover information about the inputs based on SEnSeI's output. The discriminator learns to identify which descriptor vectors were present in the input, whilst the estimator predicts the reflectance values associated with each descriptor. Both branches are trained using the mean square error function. When used with a convolutional neural network, SEnSeI takes images, rather than single reflectance values, however during unsupervised training single values are preferable, as they allow for larger batch sizes.}\label{fig:sensei-train}
\end{figure}

Autoencoders are commonly used to train models in an unsupervised manner, as we do here. The autoencoder setup involves \textit{encoding} the inputs with SEnSeI, and then \textit{decoding} those encoded features with a decoder. In our case, there were two, parallel decoder modules.  At this point, SEnSeI is not being trained for cloud masking, but simply its ability to represent the data from different spectral bands in a coherent manner. The autoencoder training setup is outlined in Fig.~\ref{fig:sensei-train}.

The first decoder module, known as the \textit{discriminator}, predicts whether or not a given reflectance band was included in the input. It outputs a confidence that a band was in the original input, doing so for twice as many candidate descriptors as were in the original, such that half its answers should be true (for those which were in the inputs), and half false (for those which were not). The loss for the discriminator, $\mathcal{L}_{discriminator}$, is the mean squared error of the predictions. The $i^{th}$ truth value is 1 if that descriptor was present in the original input, making its contribution to the loss $(1-p_i)^2$ where $p_i$ is the prediction. For fake descriptors which weren't in the original, the truth value is 0, making the term in the loss $p_i^2$. These two cases lead to the function for $\mathcal{L}_{discriminator}$ seen in Fig.~\ref{fig:sensei-train}.

Second, the \textit{estimator} takes the inputted descriptors, and determines what the reflectance value of each band was, based on the output of SEnSeI. These two different decoders lead SEnSeI to be optimized both in its ability to represent many spectral bands simultaneously without overlap and confusion, and to retain information about the reflectance values in each of those bands. The loss for this branch, $\mathcal{L}_{estimator}$, is also the mean square error. This is calculated as the square of the difference between each band's true reflectance, and the predicted reflectance outputted by the estimator (Fig.~\ref{fig:sensei-train}).

As input, SEnSeI was given a set of between 3 and 14 randomly generated descriptor vectors, alongside randomly generated reflectance values associated to each one. The descriptor vectors were created so that they represented realistic spectral bands: central wavelengths between 400nm--20$\mu$m, covering the range of typical multispectral instrument bands (see Fig.~\ref{fig:spectra}), and bandwidths also tuned to cover typical cases. Having outputted a combined feature vector, the two decoder modules (simple neural networks with several fully connected layers) performed their discrimination and estimation. Once trained, the discriminator and estimator are no longer needed and are discarded, leaving only SEnSeI to be used in the supervised training of the cloud masking model. 

The specific architecture of SEnSeI was chosen during this unsupervised training, by minimizing the loss with respect to the number of layers in each block, number of hidden units in each layer, types of pooling operation, and so on. However, like many neural network architectures, performance was not hugely sensitive to changes in these hyperparameters, and so the final version of SEnSeI was chosen from a wide set of possible configurations with similar performance. Fig.~\ref{fig:sensei} shows that architecture, as selected by this unsupervised training.

\subsection{DeepLabv3+}

DeepLabv3+~\cite{Chen2018deeplab} is a CNN designed for image segmentation, building on prior efforts~\cite{Chen2017deeplab, Chen2017rethinking}. DeepLabv3+ uses a portion of a large CNN (pre-trained on other tasks) as a feature extractor. Image segmentation performance often benefits from fusing information from different spatial scales, from the single pixel, to the whole image. DeepLabv3+ achieves this through \textit{atrous convolutions}. Atrous convolutions are like a standard convolution, but with a dilated kernel. The rate of dilation can then lead to features being extracted at different scales, by taking information from wider and wider fields of view. The resulting feature maps are combined using spatial pyramid pooling~\cite{Lazebnik2006spatial}. 

Whilst class predictions are required at every pixel, many neighbouring predictions are highly correlated with one another. Therefore, DeepLabv3+ outputs downsampled predictions, at a rate known as the \textit{output stride}, and then upsamples the predictions to the full image resolution through a bilinear interpolation. The output stride is a power of two, and is often set to 8, 16, or 32. In our work, we use an output stride of 8, given that cloud masking often requires very small areas to be segmented, unlike in other image segmentation domains where the output mask is less spatially complex. This relatively low output stride increases the computational requirements in training. For this reason, we use a relatively computationally cheap CNN as our feature extractor; MobileNetv2~\cite{Sandler2018mobnet}, which enabled us to use a higher batch size in training than would be possible with larger CNNs. Even with this smaller CNN, DeepLabv3+ has 10 times the number of trainable parameters as SEnSeI (see Appendix~\ref{appendix:computation}). Overfitting is a common problem when the number of parameters is large, but none of the models exhibited diverging validation and training losses during training, suggesting it was not a problem for them, with or without SEnSeI.

Our implementation of DeepLabv3+ uses \textit{TensorFlow 2}~\cite{Abadi2016tensorflow} and is based heavily on~\cite{Golbstein2019deeplab}. We alter the first convolutional layer of the model to accept inputs with more than the standard 3 RGB channels---64 when used with SEnSeI, or the number of bands on a satellite if used without SEnSeI.

Supervised training of our DeepLabv3+ models remained identical across experiments, other than whether or not SEnSeI was included as a preprocessing module, and which datasets were used. We used a desktop machine with an NVidia RTX 2080 Super (8GB) GPU. The models were initialized with pre-trained weights in SEnSeI, derived using the unsupervised training described in Section \ref{subsec:sens-training}, and MobileNetv2 with weights from a pre-trained version of DeepLabv3+ on the PASCAL VOC training set~\cite{Hoiem2009pascal}. All images were preprocessed according to Section \ref{subsec:preproc}, and entered the model at 257x257 pixels across, with a batch size of 8. The model was trained using stochastic gradient descent with an initial learning rate of $1\mathrm{e}{-3}$ and a momentum of 0.99. The initial learning rate was reduced by a factor of 4 when the loss converged, down to a minimum learning rate of $2\mathrm{e}{-5}$. Each epoch was 1000 steps, with a maximum of 200 epochs (although most models were stopped after around 100 epochs as they had converged on a stable loss rate). Training of the models took several hours, with each epoch taking roughly 5 minutes to complete. Models with SEnSeI took around 5\% longer per epoch, but also tended to take 10--20 more epochs to fully converge.

\subsection{Datasets}\label{subsec:data}

\begin{table*}[!t]
\centering
\caption{Overview of the labelled datasets used in our work. Whilst our own Sentinel-2 dataset has data from the most scenes, the Landsat 7 CCA and Landsat 8 CCA datasets cover many more pixels, as they use whole scenes. No-data pixels are not included in these counts. Landsat 8 SPARCS is something of an outlier regarding cloud cover percentage, with much less than other datasets.}\label{tab:datasets}
\begin{tabular}{ccccc}
\hline
\textbf{Dataset}                 &\textbf{ No. of images} & \textbf{No. of pixels (\textbf{millions})} & \textbf{Surface area (km\textsuperscript{2})} & \textbf{Cloud cover} \\ \hline
Sentinel-2 Cloud Mask Catalogue~\cite{Francis2020dataset} & 513  & 535  & 214'000 & 53.05\%  \\ \hline
Landsat 8 SPARCS ~\cite{USGS2016SPARCS} & 80  & 80 & 72'000 & 19.37\%  \\ \hline
Landsat 8 Cloud Cover Assessment~\cite{USGS2016Biome} & 96 & 4'000  & 3'600'000  &  47.91\%  \\ \hline
Landsat 7 Cloud Cover Assessment~\cite{USGS2016L7} & 197 & 7'560  & 6'800'000  &  33.23\%  \\ \hline
CloudPeru2 ~\cite{Morales2019dataset} & 153 & 5'700 & 44'400  &  48.88\%  \\ \hline
\end{tabular}
\end{table*}

\subsubsection{Sentinel-2 Cloud Mask Catalogue}
\label{subsubsec:S2dataset}

We produced a dataset~\cite{Francis2020dataset} comprising 513 subscenes taken from the 2018 archive of Sentinel-2 Level-1C imagery. Sampling of the scenes was completely random, so that the data approximated the real-world distribution of Sentinel-2 data. Each subscene and associated cloud mask is 1022 pixels squared, at 20 m resolution for all bands. In order to label a large amount of data, we designed the semi-automated tool called IRIS (Intelligent Reinforcement for Image Segmentation), available on GitHub~\cite{Mrziglod2020iris}. This tool uses a Random Forest model to extrapolate the user's paintbrush annotations across the whole image, whilst still allowing the user to make follow-up manual corrections and adjustments where needed after each run of the Random Forest.

Alongside each pixel-wise segmentation mask, each image was given several classification tags, describing the surface type, cloud thickness, cloud structure, relative cloud height, and cloud type. These tags could then be used to filter the test results, allowing us to measure performance over a range of different conditions:

\begin{enumerate}[align=left]

\item [\textbf{Surface type:}] {\textit{forest/jungle, snow/ice, agricultural, urban/developed, coastal, hills/mountains, desert/barren, shrublands/plains, wetland/bog/marsh, open\_water, enclosed\_water}}
\item [\textbf{Relative cloud height:}] {\textit{low, high}}
\item [\textbf{Cloud extent:}] {\textit{isolated, extended}}
\item [\textbf{Cloud type:}] {\textit{cumulus, cumulonimbus, stratocumulus, cirrus, haze/fog, ice\_clouds, contrails}}

\end{enumerate}

The first two authors processed the individual cloud masks, including a set of 50 which were  marked by both individuals. By comparing the two annotators' masks, we can estimate an inter-annotator agreement score, which informs users of the data what sort of accuracy their model should achieve if human-level performance is reached. The total agreement between the annotators was 95.0$\%$. The dataset is publicly available, and more information can be found in its documentation~\cite{Francis2020dataset}. In our experiments, we split the dataset in a 40:10:50 ratio, for training, validation and testing respectively.

\subsubsection{Landsat 8 SPARCS} 
\label{subsubsec:L8SPARCS}
The Spatial Procedures for Automated Removal of Clouds and Shadows (SPARCS) validation dataset~\cite{USGS2016SPARCS} consists of 80 manually created masks, each over a 1000 pixel squared subscene taken from a Landsat 8 product. As well as cloud and cloud shadow, the masks differentiate between land, water, snow/ice, and flooded surfaces. We combine these different surface types in the mask, to make the labels simply clear vs. cloud.

SPARCS is distributed in GeoTIFF format, with all bands and the mask sampled at 30m. The dataset does not contain Landsat's Panchromatic band, but includes all other 10 bands from Landsat 8. Of the 80 scenes, many show complex and varied cloud structures, suggesting the images were hand-picked. The lack of entirely cloudy or clear scenes means the dataset is not representative of real-world use on the entire Landsat 8 catalogue, but provides a challenging and diverse benchmark for models to be tested on.

\subsubsection{Landsat 8 Cloud Cover Assessment}
\label{subsubsec:L8CCA}
The USGS Landsat 8 Cloud Cover Assessment (CCA) dataset~\cite{USGS2016Biome} is designed for validation and inter-comparison of cloud masking algorithms~\cite{Foga2017Biome}. In total, 96 full Landsat 8 scenes were annotated by hand. The scenes are sampled so as to get an equal number of examples from 8 different surface types (referred to as biomes), with 12 samples each. These are; \textit{Barren, Forest, Grass/Crops, Shrubland, Snow/Ice, Urban, Water, Wetlands}. As well as annotating clear vs. cloud vs. cloud shadow, they also differentiate thick from thin cloud, however these different classes are merged in this work, in order to make it compatible with our other datasets.

\subsubsection{Landsat 7 Cloud Cover Assessment}
\label{subsubsec:L7CCA}
This dataset~\cite{USGS2016L7}, also produced by the USGS, follows a similar format to the Landsat 8 Cloud Cover Assessment dataset. 206 scenes are included, but are grouped by latitude, rather than by surface type. The classes used in annotation are also the same as those used for the Landsat 8 Cloud Cover Assessment, however the percentage of scenes annotated with cloud shadow is somewhat lower than the Landsat 8 dataset. Additionally, we found through visual inspection that some thin clouds were not marked, compared to similarly thin clouds that were marked in other datasets. During preprocessing of the dataset, we were unable to recover 9 of the 206 scenes, leaving a total of 197 for our experiments.

\subsubsection{CloudPeru2}
\label{subsubsec:CloudPeru2}

Per\'uSat-1 is a cubesat mission launched by the Peruvian space agency (CONIDA). The sensor samples data at 2.8 m resolution, in 4 spectral bands; red, green, blue and NIR. The CloudPeru2 dataset~\cite{Morales2019dataset} comprises 22'000 images, each 512x512 pixels across, taken from 153 scenes over Peru. Alongside each image is a hand-drawn mask with cloud vs. non-cloud pixels. This dataset has been used to train and test cloud masking algorithms before~\cite{Morales2019cloud}, although in our experiments we treat it only as a test set, and do not use it for training. This dataset---and those discussed previously---are summarised in Table \ref{tab:datasets}.

\subsection{Preprocessing and Formatting} \label{subsec:preproc}

For training, each image is split into 263-by-263 cropped patches and saved. Whilst DeepLabv3+ was configured to use 257-by-257 images as input, we made the crops slightly larger to introduce some variation to the samples through random translation, whilst still being small enough to be read in quickly during training. All reflectance values were converted from their native ranges to double precision floating point values between 0--1, although some pixels have reflectances higher than 1, if there is specular reflection, or if the surface is at an angle which reflects more than the normal plane could. 

For the thermal bands of Landsat 7 and 8, we converted the radiances to Brightness Temperatures (BTs), using (\ref{eqn:rad2BT}) and the parameters $K_{1}$ and $K_{2}$ in their product specifications. Then, we normalised these BTs between 0 and 1, such that a value of 1 corresponded to the 95th percentile of the total available data (\ref{eqn:BTnorm}). This was done in order to create a similar distribution of values between thermal and non-thermal bands. 

\begin{equation} \label{eqn:rad2BT}
BT = \dfrac{K_2}{ln(\dfrac{K_1}{Radiance}+1)}
\end{equation}

\begin{equation} \label{eqn:BTnorm}
BT_{normalised} = \dfrac{BT-BT_{min}}{BT_{95\%}-BT_{min}} 
\end{equation}

Alongside each image and mask pair, we also saved the wavelengths of the spectral bands held in the image, to be used as the descriptor vectors which SEnSeI takes as inputs. This was simply an array with the lower, central, and upper wavelengths of each band in nanometres, as defined in their product specifications. For input into SEnSeI, we normalise these from nanometres to a logarithmic scale, as the linear differences between e.g. red and green are far smaller than those in the SWIR or TIR range. 

\begin{equation}\label{eqn:descriptor_norm}
\lambda_{normalised} = log_{10}(\lambda - 300) - 2
\end{equation}

During training, data augmentation was applied as it has been shown to help prevent over-fitting, and improve model performance~\cite{van2001augmentation}. We included random flipping, rotation, translation, addition of white noise, salt-and-pepper noise, and small random shifts in overall reflectances across the image's bands.

\subsection{Performance Metrics} \label{subsec:metrics}

Trying to summarise the performance of models in an image segmentation task with a single number is problematic. Whilst it can be useful to judge all models with a single variable, this obscures the reality that models perform in different ways, with different strengths and weaknesses. The experiments in this study use five well-known metrics so that a more nuanced account of the models' performance can be made.

All five metrics are defined by the populations of True Positive (correctly identified cloud), True Negative (correctly identified non-cloud), False Positive (non-cloud labelled incorrectly as cloud) and False Negative (cloud incorrectly labelled as non-cloud).

Overall Accuracy (OA) is the total percentage of correctly identified pixels. It is an intuitive metric, but can be unrepresentative of performance when class prevalences are very imbalanced:

\begin{equation}
OA = 100*\frac{TP+TN}{TP+TN+FP+FN}
\end{equation}

Recall tells us how successful the model is in finding the population of cloudy pixels. It does not take into account the fact that the model might incorrectly label pixels as cloud (False Positive):

\begin{equation}
Recall = 100*\frac{TP}{TP+FN}
\end{equation}

Precision, as opposed to recall, determines the percentage of pixels which were identified as cloudy that are correct. It does not depend on the number of missed cloudy pixels (False Negatives):

\begin{equation}
Precision = 100*\frac{TP}{TP+FP}
\end{equation}

The $F_1$ score is the harmonic average between precision and recall. This means that it always lies between their values, but it is heavily penalised if one is much lower than the other. Of the metrics commonly found in deep learning, $F_1$ is one of the most widely used as a singular measure of a model's performance, given that it takes into account both precision and recall. However, the $F_1$ score still does not take into account the True Negative population, which can lead to very low scores when the negative class is much more common than the positive one:

\begin{equation}
F_1 = 100*\frac{2*Precision*Recall}{Precision+Recall}
\end{equation}

Balanced Accuracy (BA) is similar to OA, but it is weighted by the relative prevalence of each class (in this case, cloud or non-cloud). It can also be thought of as the average of each class' recall:

\begin{equation}
BA = 100*0.5*\left(\frac{TP}{TP+FN} + \frac{TN}{TN+FP}\right)
\end{equation}

\section{Experimental Results}\label{sec:results}

\subsection{Sentinel-2} \label{subsec:exp-S2}

This experiment was conducted to measure the performance of both the standard DeepLabv3+ model, and the sensor independent variants with SEnSeI prepending DeepLabv3+, alongside some other cloud masking options that are widely used. The three main goals of this experiment were:

\begin{enumerate}
\item{Demonstrate a state-of-the-art cloud masking model for Sentinel-2.}
\item{Investigate whether Hyp. (A) and Hyp. (B) from Section \ref{sec:intro} would hold, or if the addition of SEnSeI, and an increasingly generalised training configuration, would cause performance to suffer against the specialised, non-sensor independent model.} 
\item{Determine how a deep learning-based approach compared to other popular Sentinel-2 cloud masking algorithms.}
\end{enumerate}

The Sentinel-2 dataset was split randomly into training, validation, and testing sets, such that 40\% (205 subscenes) were in training, 10\% (51 subscenes) in validation, and 50\% (257 subscenes) in testing. The training for each model was then carried out as Section \ref{subsec:preproc} details. 

Two versions of DeepLabv3+ were trained without SEnSeI. First, a version simply trained with Sentinel-2 data, using all bands (referred to as `\textit{DLv3 - S2}' in the experiments). Second, a version trained on both the Sentinel-2 training data, and the Landsat 8 CCA dataset, using only  those bands which are found on both Sentinel-2 and Landsat 8, making it an example of the \textit{transfer learning} approach outlined in Section \ref{subsec:cloudmasking}. This model is referred to as `\textit{DLv3 - S2\&L8 [common bands]}'. The common bands---in the naming scheme of Sentinel-2---were bands B1, B2, B3, B4, B8, B10, B11, and B12.

\begin{table}[]
\renewcommand{\arraystretch}{1.7}
\setlength{\tabcolsep}{3.5pt}
\centering
\caption{The five performance metrics for each model, across the 257 Sentinel-2 subscenes used as a test set. The highest performance for each metric is highlighted in bold. All variants of DeepLabv3+ outperform the other models in OA, BA, Recall and $F_1$, however s2cloudless does achieve the highest precision.}\label{tab:exp-S2}
\begin{tabular}{c|ccccc}
\textbf{Model} & \textbf{OA} & \textbf{BA} & \textbf{Prec.} & \textbf{Rec.}  & $\mathbf{F_1}$ \\ \hline
DLv3 - S2                          & 94.04 & 94.03 & 94.57 & 94.17 & 94.37 \\
DLv3 - S2\&L8 {[}common bands{]}   & \textbf{95.19} & \textbf{95.18} & 95.61 & \textbf{95.31} & \textbf{95.46} \\
SEnSeI+DLv3 - S2 {[}fixed bands{]} & 94.46 & 94.42 & 94.58 & 94.99 & 94.79 \\
SEnSeI+DLv3 - S2                   & 93.08 & 93.10 & 94.00 & 92.89 & 93.44 \\
SEnSeI+DLv3 - S2\&L8               & 93.73 & 93.73 & 94.42 & 93.73 & 94.07 \\
\makecell{SEnSeI+DLv3 - S2\&L8 \\ {[}no pre-training of SEnSeI{]}} & 93.11 & 93.16 & 94.47 & 92.43 & 93.44 \\ \hline
NN - S2 						 	 & 90.04 & 90.04 & 91.06 & 90.08 & 90.57 \\
NN - S2\&L8 {[}common bands{]}	 & 89.86 & 89.98 & 92.50 & 88.02 & 90.20 \\
SEnSeI+NN - S2				 	 & 88.90 & 88.75 & 88.19 & 91.31 & 89.72 \\
SEnSeI+NN - S2\&L8			 	 & 89.04 & 89.16 & 91.70 & 87.23 & 89.41 \\ \hline
CloudFCN - S2 \cite{Francis2019cloudfcn} & 91.58 & 91.67 & 93.58 & 90.34 &91.93 \\
s2cloudless \cite{Zupanc2017s2cloudless}   & 92.42 & 92.61 & \textbf{95.90} & 89.54 & 92.61 \\
Standard L1C product mask \cite{Coluzzi2018L1C} & 83.11 & 83.29 & 86.88 & 80.27 & 83.45
\end{tabular}
\end{table}

Four different configurations of DeepLabv3+ were used for SEnSeI. All four of these had the exact same architecture. Three began training with the same pre-trained weights in both SEnSeI and DeepLabv3+. These three configurations are increasingly sensor independent, adding to the ability of the model to generalise to other sensors. Another configuration used a SEnSeI with randomly initialized weights, to see what advantage pre-training offered:

\begin{enumerate}
\item{`\textit{\textbf{SEnSeI+DLv3 - S2 [fixed bands]}}' was given all Sentinel-2 bands at every training step. This means it was only ever given Sentinel-2 data with all bands present, thus producing a model that could not be generalised to work on any other satellite or subset of Sentinel-2 bands, but containing the SEnSeI architecture.}

\item{`\textit{\textbf{SEnSeI+DLv3 - S2}}', was trained using random combinations of the available Sentinel-2 bands, from  a minimum of 3, up to a maximum of all 13. This additional complexity in training produces a model which is theoretically capable of predicting clouds from sensors with any combination of the existing Sentinel-2 bands (this assumption of sensor independence will be tested further in Section \ref{subsec:exp-misc}).}

\item{`\textit{\textbf{SEnSeI+DLv3 - S2\&L8}}', was also given the Landsat 8 Cloud Cover Assessment dataset as training data. It then received random combinations of the available bands on each satellite, creating what is shown in later experiments to be a truly sensor independent model.}

\item{`\textit{\textbf{SEnSeI+DLv3 - S2\&L8 [no pre-training of SEnSeI]}}', was the same as the previous but the weights in SEnSeI were randomly initialized.}

\end{enumerate}

As well as DeepLabv3+, we combine SEnSeI with a 2-layer neural network, which has no convolutions. This neural network has only a few hundred parameters, and represents an extremely simple machine learning solution. By combining SEnSeI with both this small neural network, and the very complex DeepLabv3+, we hope to show that SEnSeI works with a wide range of model architectures. The naming conventions for the neural network model configurations are the same as for DeepLabv3+, with ``NN'' replacing ``DLv3''.

In addition to DeepLabv3+ and the neural network, we also attempted to train CloudFCN~\cite{Francis2019cloudfcn} with SEnSeI to further show SEnSeI's general applicability; however the training became unstable when they were used together, and it was not possible to get the models to reliably converge. The reason for this is not clear, but it may be that, when used with SEnSeI, CloudFCN would need to be trained with a larger batch size than we were able to process on the 8GB of video memory used in training. This is an avenue for future research, as it may be that SEnSeI adds some instability to certain model architectures (perhaps those with an architecture similar to U-Net~\cite{Ronneberger2015unet}, like CloudFCN) during training. We do however include results from CloudFCN without SEnSeI in this experiment, trained on the Sentinel-2 dataset.

We also include results from \textit{s2cloudless}~\cite{Zupanc2017s2cloudless}, using a confidence threshold of 0.5. This algorithm is a pixel-based machine learning model, which uses the XGBoost algorithm~\cite{Chen2016xgboost}. 

Lastly, the \textit{Standard L1C product mask} is the cloud mask generated during L1C processing~\cite{Coluzzi2018L1C}. We create a rasterized version of the polygonal information in the GML file provided with each L1C product. All ``OPAQUE'' and ``CIRRUS'' polygons were included. The full product masks were then cropped to the extent of the subscene, and visually inspected to confirm they had been correctly processed and were aligned with the image. Comparison with this mask allows us to judge the relative performance of a deep learning-based model to a thresholding model.

\begin{figure*}
\centering
\includegraphics[width=0.825\linewidth , trim=0cm 0.33cm 0cm 0cm , clip]{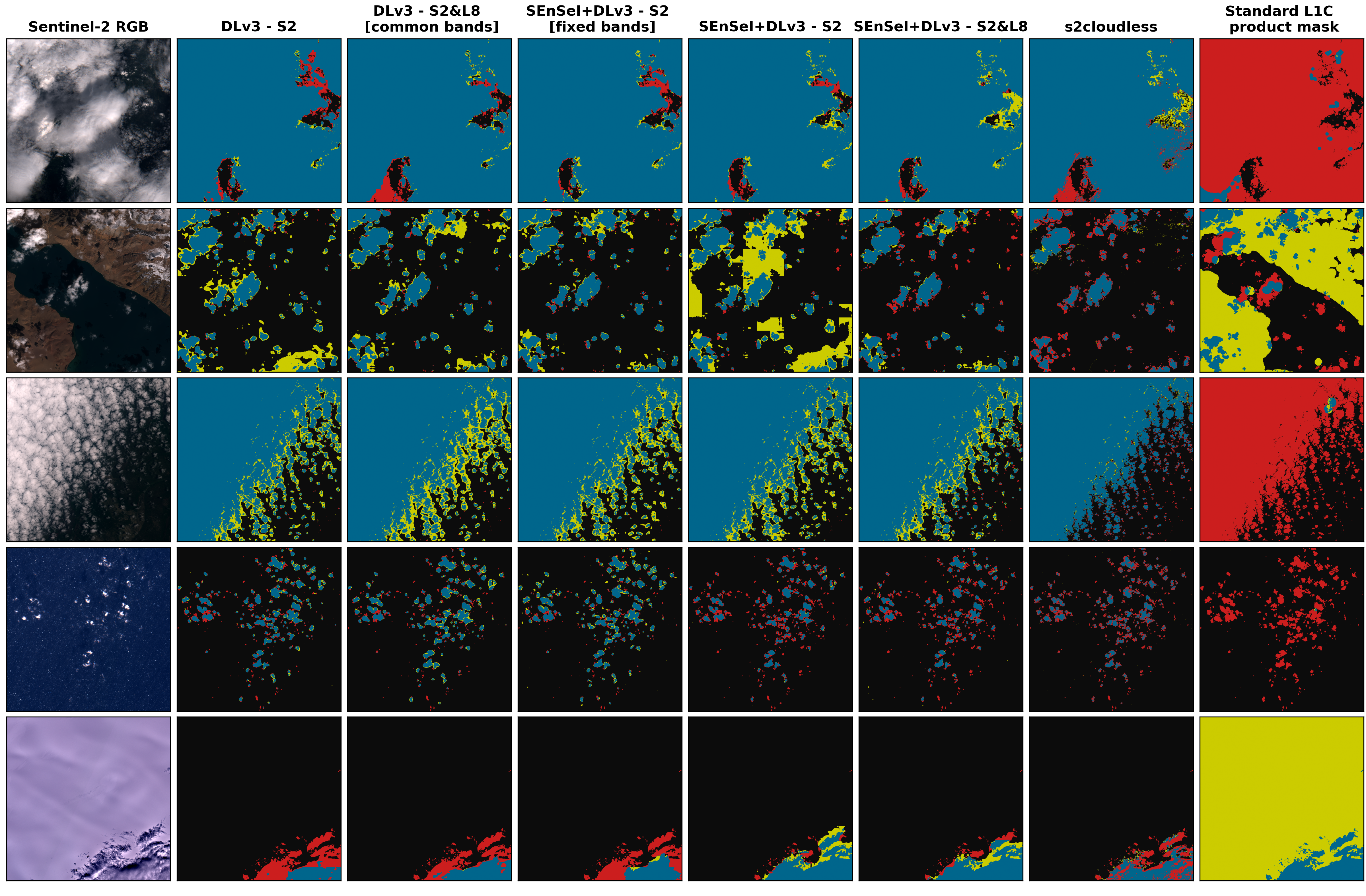}
\caption{Visualisation of seven models' predictions tested on five scenes from the Sentinel-2 test set. Blue is correctly identified cloud, red is missed cloud, and yellow is false cloud detection. These scenes were selected as they contained interesting, diverse and challenging cloud structures.}
\label{fig:exp-1-vis}
\end{figure*}

To assess each model's performance in this experiment, we consider five metrics. Overall Accuracy (\textit{OA}), Balanced Accuracy (\textit{BA}), Precision, Recall, and $F_{1}$, as defined in Section \ref{subsec:metrics}. These are calculated across the entire Sentinel-2 test set. Table \ref{tab:exp-S2} provides numerical values for these metrics over the dataset. Fig. \ref{fig:exp-1-vis} gives some visual examples of the masks produced by the models. Appendix \ref{appendix:results} gives a further breakdown of results for the models across different image types.

Overall, \textit{DLv3 - S2\&L8 [common bands]} performs better than the other models. This shows that transfer learning techniques (as used by others, e.g.~\cite{Wieland2019multi, Shendryk2019deep,Li2019deep}) can result in boosts to performance, despite the removal of the bands which Landsat 8 does not also include. Besides this model, \textit{SEnSeI+DLv3 - S2 [fixed bands]} also exceeds the performance of \textit{DLv3 - S2} in all metrics, showing that prepending SEnSeI to a deep learning model does not produce a significant drop in performance, supporting Hyp. (A). When a more challenging training configuration was used, in which band combinations were sampled randomly, performance does drop somewhat (\textit{SEnSeI+DLv3 - S2}). Promisingly, however, performance of \textit{SEnSeI+DLv3 - S2\&L8} is higher than that of \textit{SEnSeI+DLv3 - S2} in every metric, making it clear that adding data from more than one satellite boosts the model's reliability. Whilst \textit{SEnSeI+DLv3 - S2\&L8} falls short of \textit{DLv3 - S2\&L8 [common bands]} in terms of numerical performance, it is important to remember that the former is able to produce cloud masks over a huge range of different spectral band combinations, whilst the latter requires all bands shared by Sentinel-2 and Landsat 8 to be present. We also observe a small decrease in performance of roughly 0.6\% in OA, BA, and $F_1$ when using a randomly initialised version of SEnSeI, rather than the pre-trained weights. This suggests pre-training SEnSeI is worthwhile, and has a moderate but noticeable impact on performance.

SEnSeI's effect on DeepLabv3+ is mirrored closely by its effect on the 2-layer neural network. The creation of a sensor independent model leads to a drop of about 1\% from \textit{NN - S2} to \textit{SEnSeI+NN - S2} in OA, BA and $F_1$. This 1\% drop in performance (on test data from Sentinel-2) seems to be the cost of making a model which works across many sensors when using SEnSeI. Interestingly, the addition of data from Landsat 8 did not boost the neural network's performance, like it did for DeepLabv3+, suggesting that such a small, simple model as this cannot benefit from an expanded training set, whereas DeepLabv3+ can.

In order to determine whether the differences in performance seen in Table \ref{tab:exp-S2} are statistically significant or not we carried out a cross-validation and bootstrapping experiment to investigate the uncertainties in our results. Due to computational limitations in training many models, we were only able to analyse the uncertainty for just two of the model configurations (\textit{DLv3 - S2} and \textit{SEnSeI+DLv3 - S2\&L8}). Although not all models were included, we believe that given their architectures, training data, and the test set are identical to those not included, this analysis can be safely extrapolated to approximate the uncertainty for all our model results. For each model configuration, we use 5-fold cross-validation to train 5 different models, each with a different 40\% of the data as training, and 10\% as validation. The same 257 scenes are always held as the test set for all models. We then bootstrap the test set results, with 1 million iterations per model, to measure the data-driven uncertainty in our performance. With this method, both models tested were found to have a range of $\pm0.5\%$  in OA, BA and $F_1$. We take this range as a rough limit for statistical significance in our analysis of the results.

\begin{figure*}
\centering
\includegraphics[width=0.825\linewidth , trim=0cm 0.33cm 0cm 0cm , clip]{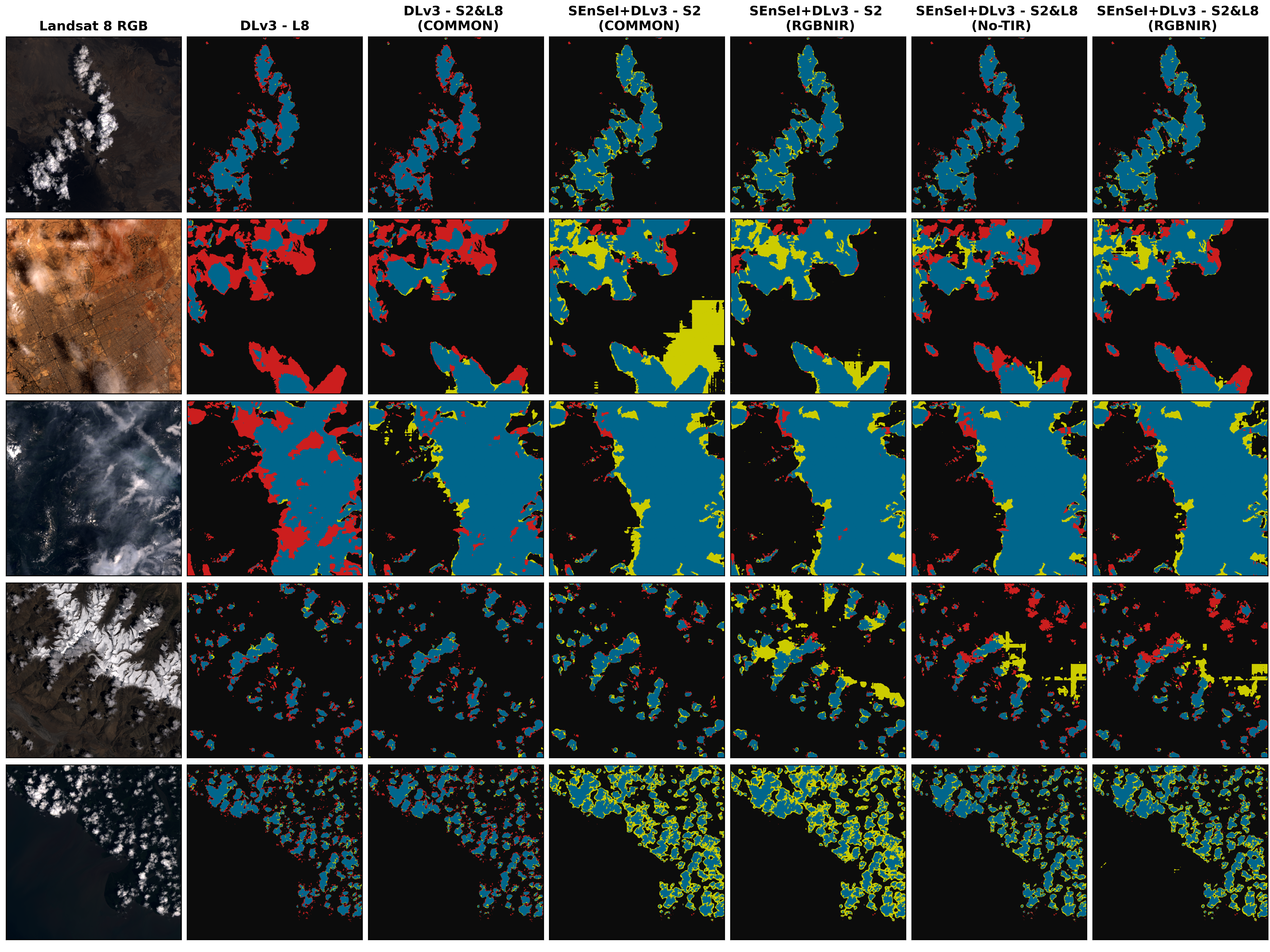}
\caption{Visualisation of six models' predictions tested on five scenes from the Landsat 8 SPARCS test set. Blue is correctly identified cloud, red is missed cloud, and yellow is false cloud detection. \textit{DLv3 - L8} shows much higher precision and lower recall than other models. The boundaries of the thin cloud in the 2nd row are understandably error-prone, given the smooth transition between cloudy and clear. Some models misclassify parts the snow-covered mountains in the 4th row as cloud, but successfully reject the snow on the left of the 3rd row's scene.}
\label{fig:L8-vis}
\end{figure*}

\subsection{Landsat 8}
\label{subsec:exp-L8}

We now present results on the Landsat 8 SPARCS dataset (described in Section \ref{subsubsec:L8SPARCS}). In testing these models, we demonstrate true sensor independence (a model working on multiple satellite sensors at once), and show strong evidence for Hyp. (B) and Hyp. (C). In addition, we also use this experiment as an opportunity to investigate the effects of different band combinations, by testing the same models with different spectral band combinations. The first model, \textit{DLv3 - L8}, is trained using the 96 scenes from the Landsat 8 Cloud Cover Assessment dataset, but without the panchromatic band (to make the model compatible with the SPARCS dataset). \textit{DLv3 - S2\&L8 [common bands]} is also used in this experiment, and is the same model as that used in Section \ref{subsec:exp-S2}. Then, in a similar fashion to Section \ref{subsec:exp-S2}, we include two configurations of DeepLabv3+ with SEnSeI:

\begin{enumerate}
\item{`\textit{\textbf{SEnSeI+DLv3 - S2}}' is the same model (with the exact same weights) as used in Section \ref{subsec:exp-S2} with mixed band combinations as input in training. No retraining takes place to prepare it for use with Landsat 8, however, bands that Sentinel-2 does not have (thermal bands) are not used in testing.}
\item{`\textit{\textbf{SEnSeI+DLv3 - S2\&L8}}' is also the identical model to that which is used in Section \ref{subsec:exp-S2}, with no retraining.}
\end{enumerate}

For both of these SEnSeI configurations, we run tests on SPARCS using several different band combinations. For 10 bands, there are 968 possible combinations of 3 or more bands. We therefore only select a few of interest:
\begin{enumerate}
\item{`\textit{\textbf{ALL}}': All Landsat 8 bands (excluding the panchromatic band, which is not distributed with the SPARCS dataset).}
\item{`\textit{\textbf{COMMON}}': All bands shared by Sentinel-2 and Landsat 8. SPARCS already excludes Landsat 8's panchromatic band, so this band combination is just all available bands except the two thermal bands.}
\item{`\textit{\textbf{RGB}}': Bands 2, 3, and 4.}
\item{`\textit{\textbf{AeroRGB}}': Bands 1, 2, 3, and 4.}
\item{`\textit{\textbf{AeroRGBNIR}}': Bands 1, 2, 3, 4 and 5.}
\item{`\textit{\textbf{No-RGB}}': All possible bands except 2, 3 and 4.}
\end{enumerate}

\begin{table}
\renewcommand{\arraystretch}{1.25}
\setlength{\tabcolsep}{2.5pt}
\centering
\caption{Results of models across the Landsat 8 SPARCS dataset. The versions of RS-Net which are highlighted were tested on SPARCS using 5-fold cross-validation, and thus have an advantage over other models in that they have been exposed to the style and distribution of SPARCS at training. Therefore, in metrics for which this version of RS-Net is best, we also highlight the best model not trained on SPARCS. A version of Cloud-Net+ was trained using 64 SPARCS images and tested on 16, but is omitted here as the results do not cover all 80 scenes.}\label{tab:exp-L8}
\begin{tabular}{c|cccccc}

\textbf{Model} & \textbf{Bands} & \textbf{OA} & \textbf{BA} &  \textbf{Precision} &  \textbf{Recall} &  $\mathbf{F_1}$ \\ \hline

DLv3-L8 & ALL & 93.03 &  83.32 & 95.17 &   67.46 &  78.96 \\  \hline
DLv3-S2\&L8 & \makecell{COMMON} & 93.70 &  87.04 & 89.82 & 76.14 & 82.42 \\  \hline

    & COMMON &  91.86 &  \textbf{93.05} &      71.96 &   \textbf{94.99} &  81.89 \\
    & RGB &  90.91 &  91.77 &      69.92 &   93.18 &  79.89 \\
    & RGBNIR &  91.11 &  92.23 &      70.18 &   94.06 &  80.38 \\
    & AeroRGBNIR &  91.11 &  92.45 &      70.02 &   94.64 	&  80.49 \\
\multirow{-5}{*}{\makecell{SEnSeI+DLv3 \\ S2}} & No-RGB & 91.52	& 92.45	& 71.34	& 93.96	& 81.10 \\ \hline

	& ALL &  93.91 &  86.31 &      93.24 &   73.90 &  82.45 \\
	& COMMON &  \textbf{94.24} &  90.98 &      84.77 &   85.65 &  \textbf{85.21} \\ 
	& RGB&  92.45 &  92.34 &      74.73 &   92.18 &  82.54 \\ 
	& RGBNIR &  93.51 &  92.29 &      79.12 &   90.31 &  84.34 \\
  & AeroRGBNIR &  93.63 &  91.67 &      80.54 &   88.47 &  84.32 \\ 
\multirow{-6}{*}{\makecell{SEnSeI+DLv3 \\ S2\&L8}}          & No-RGB &  93.83 &  86.19 &      93.00 &   73.71 &  82.24 \\  \hline

 & ALL & 92.53 & -- &88.57 & 70.53 & 78.35 \\
 & COMMON & 93.26 & -- & 91.04 & 72.34 & 80.62 \\
 & RGBNIR & 92.53 & -- & \textbf{95.35} & 64.56 & 76.99 \\
\multirow{-4}{*}{\makecell{RS-Net \\ trained \\ on Biome \\ \cite{Jeppesen2019rsnet}}} & RGB & 92.38 & -- & 86.54 & 71.83 & 78.50 \\ \hline

\rowcolor{lightred} & ALL & 94.54 & -- & 87.62 & 83.66 & 85.59 \\
\rowcolor{lightred} & COMMON & \textbf{\textit{95.60}} & -- & 89.47 & 87.58 & 85.59 \\
\rowcolor{lightred} & RGBNIR & 94.85 & -- & 88.92 & 83.89 & 86.33 \\
\rowcolor{lightred} \multirow{-4}{*}{\makecell{RS-Net \\ trained \\ on SPARCS \\ (5-fold C.V.) \\ \cite{Jeppesen2019rsnet}}} & RGB & 94.86 & -- & 88.58 & 84.34 & \textbf{\textit{86.41}} \\ \hline

\makecell{Cloud-Net+ \\ trained  on \\ 38-Cloud \cite{Mohajerani2020cloud}} & RGBNIR & 91.30 & 86.47 & 76.94 & 78.60 & 77.76 \\ \hline

\makecell{Cloud-Net+ \\ trained on \\ 95-Cloud \cite{Mohajerani2020cloud}} & RGBNIR & 90.45 & 85.69 & 74.14 & 77.90 & 75.97 \\ \hline

Fmask \cite{Jeppesen2019rsnet} & -- & 92.47 & -- & 77.47 & 86.21 & 81.61 \\

\end{tabular}

\end{table}

As well as our own models' results, we also include the results of Fmask, the standard Landsat 8 cloud masking software, and RS-Net, another state-of-the-art deep learning model discussed in Section \ref{subsec:cloudmasking}, which was trained on several different band combinations of Landsat 8. We also compare the results of Cloud-Net+ when trained on their 38-Cloud and 95-Cloud datasets and tested on SPARCS. We did not compute predictions for these three models ourselves, but simply recover the results from prior studies (\cite{Jeppesen2019rsnet} for Fmask and RS-Net,~\cite{Mohajerani2020cloud} for Cloud-Net+). For Fmask, there are differing results on the SPARCS dataset in different studies, possibly because different versions of the code of were used, and so the best published result for the SPARCS dataset is taken~\cite{Jeppesen2019rsnet}.

The results of this experiment are shown in Table \ref{tab:exp-L8}, and some of the models can be compared visually in Fig. \ref{fig:L8-vis}. Out of the models not trained on SPARCS data, \textit{SEnSeI+DLv3 - S2\&L8 (COMMON)} achieves the highest OA and $F_1$ score, although it is beaten by models trained using 5-fold cross-validation on SPARCS. Perhaps a more important result for Hyp. (C), though, is the performance of \textit{SEnSeI+DLv3 - S2} across the different band combinations, which shows a model performing with $>90\%$ OA on data from a sensor it has not seen in training. Falling just short of Fmask in OA, we believe this shows adequate performance for a model working on a different sensor. Finally, the significant increase in performance between \textit{DLv3 - L8} and \textit{SEnSeI+DLv3 - S2\&L8}, shows that adding data from another satellite (Sentinel-2) \textit{increases} performance on Landsat 8, agreeing with Hyp. (B). Not only is SENSeI allowing for models to become more generalised and work across different sensors, but, in this instance, it is allowing models to enhance their predictive capabilities on one sensor with data from another.

The change in performance of the same models with different band combinations is relatively small. For example, OA for most models varies by around 1-2\%, whilst the $F_1$ score is slightly more sensitive, varying by around 2-3\%. Generally, models with limited band combinations, e.g. RGB and RGBNIR, suffered in comparison to more extensive band combinations, which makes sense given that they do not have access to as much spectral information from their inputs. Surprisingly, for all three models which were tested both with and without the thermal infrared bands (\textit{SEnSeI+DLv3 - S2\&L8}, and both versions of RS-Net), performance actually \textit{decreased} when they were included. This is surprising, given that thermal bands provide quite different information to the others, which would ordinarily be a good thing for a model to have.

\subsection{Other Satellites}
\label{subsec:exp-misc}

In this section we seek to test the limits of SEnSeI's sensor independence, by predicting cloud masks for satellites not included in the model's training, and different resolutions to the 20--30 m/pixel range of Sentinel-2 and Landsat 8. We include three sensors in the following experiments, two of which (Landsat 7 and Per\'uSat-1) have datasets that allowed us to measure our models' performance, and another (Sentinel-3 SLSTR) which we include for visual inspection only, as no suitable dataset was available.

Landsat 7's non-thermal bands all have close equivalents in Sentinel-2 and Landsat 8. It's thermal band, whilst not directly shared by Landsat 8, spans roughly the equivalent of both of Landsat 8 TIR bands, and so can be approximately simulated by averaging them together. Whilst the spectral profiles and resolution of Landsat 7 are familiar to our model, the sensor design and image quality differ between Landsat 7 and the other satellites. An 8-bit digitisation depth was used for Landsat 7, unlike the larger bit depths on more recent missions, which meant that some of the higher reflectances are saturated, in order to allow for more radiometric precision at lower reflectances. This leads some high reflectance surfaces such as snow, ice and clouds, to have a textureless appearance, as many pixels are at the maximum of the range.

\begin{figure*}
\centering
\includegraphics[width=0.83\linewidth]{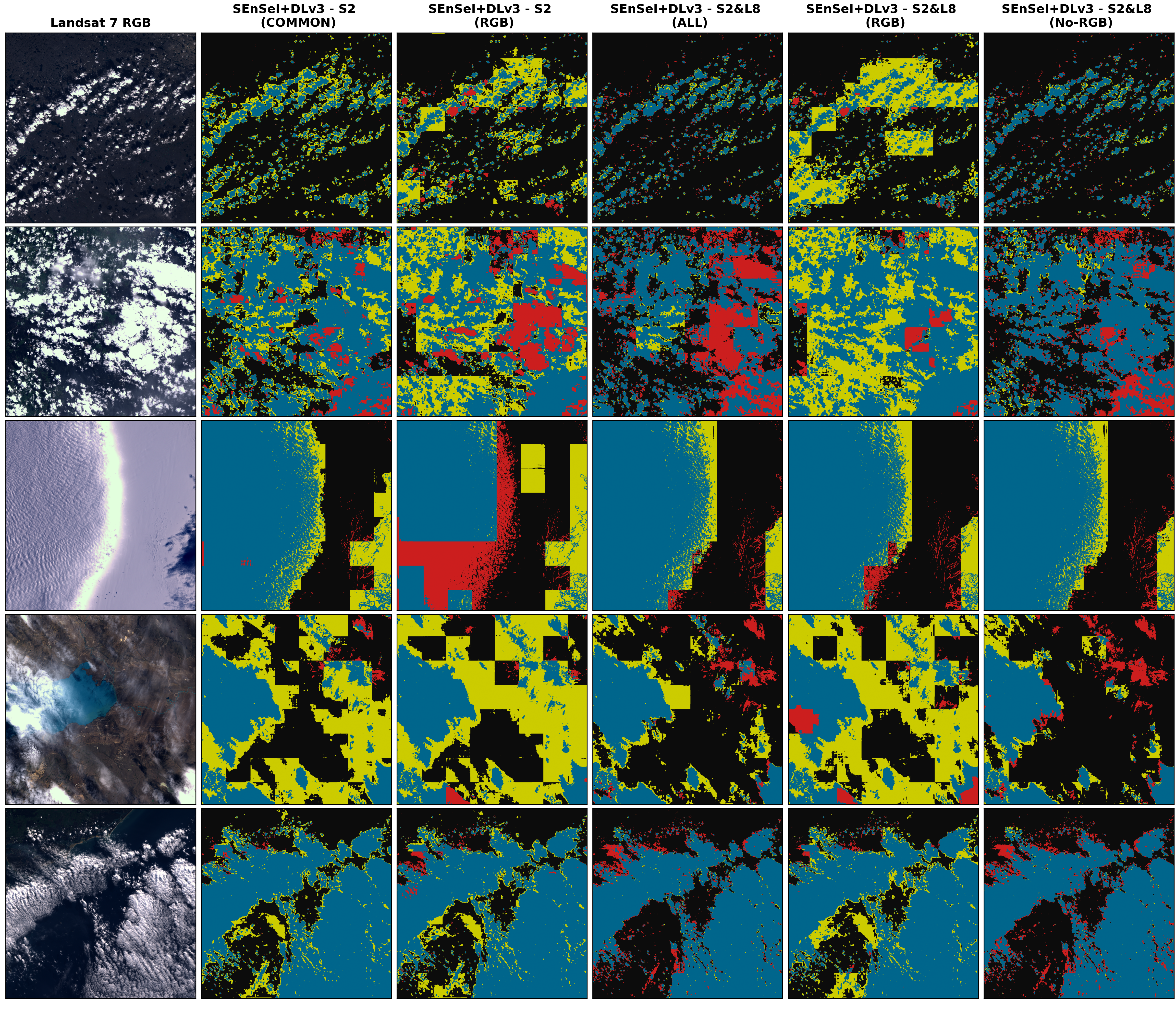}
\caption{Visualisation of five models' predictions tested on 6 scenes from the Landsat 7 CCA test set. We take 2000 pixel squares from the scenes, as the full scenes are too detailed to display in this manner. Models using RGB bands are inconsistent, while those using other bands are more successful. The fourth row shows an image with very thin cloud which is picked up by some of the models, but is not annotated as cloud.}
\label{fig:L7-vis}
\end{figure*}

We use the large Landsat 7 CCA dataset (described further in Section \ref{subsubsec:L7CCA}) to test SEnSeI's performance. As before, we use versions of SEnSeI trained either on Sentinel-2, or Sentinel-2 and Landsat 8, and compare the test set results with different band combinations as input. We compare our results to the those published in 2012~\cite{Scaramuzza2011CCA}, which present accuracy scores for three models over 103 of the Landsat 7 CCA images. As well as the numerical findings in Table \ref{tab:exp-L7}, we also visualise of some failures and successes, in Fig. \ref{fig:L7-vis}.

\begin{table}
\renewcommand{\arraystretch}{1.5}
\setlength{\tabcolsep}{2.5pt}

\caption{Results for models over the 197 scenes from the Landsat 7 CCA dataset. Models using only RGB made substantial mistakes, perhaps due to saturation in Landsat 7 images. The combination of all bands other than RGB (No-RGB) were generally the best, and the model trained on both Sentinel-2 and Landsat 8 data is generally better than the Sentinel-2-only model.}\label{tab:exp-L7}

\begin{tabular}{c|cccccc}
                                             Model & Bands & OA & BA &  Precision &  Recall &     F1 \\ \hline
                                             
\multirow{3}{*}{\makecell{SEnSeI+DLv3 \\ S2}} & RGB &  76.23 &  73.17 &  64.27 &   64.05 &  64.16 \\
      & COMMON &  85.24 &  86.25 &      72.61 &   89.26 &  80.08 \\
      & No-RGB &  85.25 &  86.63 &      72.09 &   \textbf{90.73} &  80.34 \\ \hline                                            
\multirow{3}{*}{\makecell{SEnSeI+DLv3 \\ S2\&L8}} 
      & RGB &  74.62 &  71.72 &      61.52 &   63.06 &  62.28 \\
      & ALL &  89.08 &  87.77 &      83.35 &   83.88 &  83.62 \\
      & No-RGB &  \textbf{89.90} &  \textbf{88.90} &  \textbf{84.03} &   85.92 &  \textbf{84.96} \\ \hline

C5 CCA \cite{Scaramuzza2011CCA} & - & 88.5 & - & - & - & - \\ \hline
AT-ACCA \cite{Scaramuzza2011CCA} & - & 76.6 & - & - & - & - \\ \hline
\makecell{Expanded \\ AT-ACCA \cite{Scaramuzza2011CCA}} & - & 89.7 & - & - & - & - \\
\end{tabular}

\end{table}

\begin{table}
\caption{Results of models across the CloudPeru2 dataset, from Per\'uSat-1. We test our two models at the original resolution (2.8 m) and 5.6 m. The masks at the coarser resolution are then bilinearly resampled to retrieve a mask at the full 2.8 m resolution. The models examined in~\cite{Morales2019cloud} are included, but direct comparison to them is not strictly valid, given that they tested the models on 5\% of the available data, using the rest for training.}\label{tab:CloudPeru2}
\renewcommand{\arraystretch}{1.5}
\setlength{\tabcolsep}{2.5pt}
\centering

\begin{tabular}{c|cccccc}
                        \textbf{Model} & \textbf{Bands} & \textbf{OA} & \textbf{BA} &  \textbf{Precision} &  \textbf{Recall} & \textbf{F1} \\
\hline
      & RGB &  90.86 &  90.90 &      89.03 &   92.71 &  90.83 \\
   \multirow{-2}{*}{\makecell{SEnSeI-DLv3 \\ S2 (2.8 m)}} & RGBNIR &  90.95 &  91.00 &      89.04 &   92.92 &  90.94 \\ \hline
   & RGB &  91.63 &  91.68 &      89.78 &   \textbf{93.52} &  91.61 \\
 \multirow{-2}{*}{\makecell{SEnSeI-DLv3 \\ S2\&L8 (2.8 m)}} & RGBNIR &  91.52 &  91.56 &      89.97 &   93.01 &  91.47 \\ \hline

      & RGB &  92.22 &  92.20 &      92.41 &   91.59 &  92.00 \\
   \multirow{-2}{*}{\makecell{SEnSeI-DLv3 \\ S2 (5.6 m)}} & RGBNIR &  92.51 &  92.50 &      92.72 &   91.88 &  92.30 \\ \hline
   & RGB &  92.70 &  92.67 &      93.49 &   91.42 &  \textbf{92.44} \\
 \multirow{-2}{*}{\makecell{SEnSeI-DLv3 \\ S2\&L8 (5.6m)}} & RGBNIR &  \textbf{92.74} &  \textbf{92.69} &     \textbf{ 94.36} &   90.54 &  92.41 \\ \hline
 \rowcolor{lightred} CloudNet \cite{Mohajerani2019cloud} (2.8 m) & RGBNIR & 94.01 & 93.91 & \textit{\textbf{97.82}} & 89.78 & 93.63 \\ \hline
 \rowcolor{lightred} \cite{Morales2017cloud}'s method (2.8 m) & RGBNIR & 91.34 & 91.97 & 86.52 & 92.36 & 89.34 \\ \hline
 \rowcolor{lightred} \cite{Morales2019cloud}'s method (2.8 m) & RGBNIR & \textit{\textbf{97.50}} & \textit{\textbf{97.52}} & 96.45 & \textit{\textbf{98.46}} & \textit{\textbf{97.44}} \\
\end{tabular}
\end{table}

Overall, the models are reasonably successful on Landsat 7, achieving accuracies close to 90\%, with the best version just out-performing the highest reported accuracy from~\cite{Scaramuzza2011CCA}. However, there is a noticeable increase in error rate when compared to our results on Sentinel-2 and Landsat 8. In particular, the model appeared to struggle in regions of the image which were uniformly high reflectance, and when it was given only RGB bands. This may have been caused by Landsat 7's tendency to saturate pixels with high reflectances, leading to a lack of texture which perhaps confused the model. It may be possible to mitigate these errors by adding artificial saturation to images in training. 

Another likely reason for reduced accuracy on Landsat 7 is the discrepancy in annotation styles between datasets, with the Landsat 7 CCA dataset classifying as clear some areas that we would judge to be thin cloud, which was also noted by~\cite{Scaramuzza2011CCA}. This means that a model trained on the Sentinel-2 and Landsat 8 datasets is biased in relation to the Landsat 7 CCA labels. We still treat this experiment as partially successful, given the challenges, but we suggest more work is needed on understanding why the model failed where it did, and why in this instance Hyp. (C) is not well supported.

Per\'uSat-1's resolution is an order of magnitude finer than that of Sentinel-2 and Landsat 8, at 2.8 m/pixel. We use the CloudPeru2 dataset (Section \ref{subsubsec:CloudPeru2}) to see how a sensor independent model copes with a significant change in scale. The model results are given in Table \ref{tab:CloudPeru2}. In order to investigate how a change in scale affects results, we produce cloud masks at both the original resolution (2.8 m/pixel) and at 50\% downsampling, with an effective resolution of 5.6 m/pixel. 

Direct comparison with the other models included in the table is difficult, as those considered by~\cite{Morales2019cloud} were all trained on 90\% of the dataset, and tested only on 5\%, giving their models a significant advantage in having been exposed to the same distribution of data as the test set. Given this difference in approach, we believe our results, especially at 5.6 m, resolution are excellent, out-performing some of the models trained specifically on the dataset. The drop in performance seen at 2.8 m is perhaps a sign that we are reaching the limit of our model's resolution range, having only been trained on a resolution of 20 m/pixel or more.

\begin{figure*}
\centering
\includegraphics[trim=0cm 3.8cm 0cm 2.5cm, clip, width=0.96\linewidth]{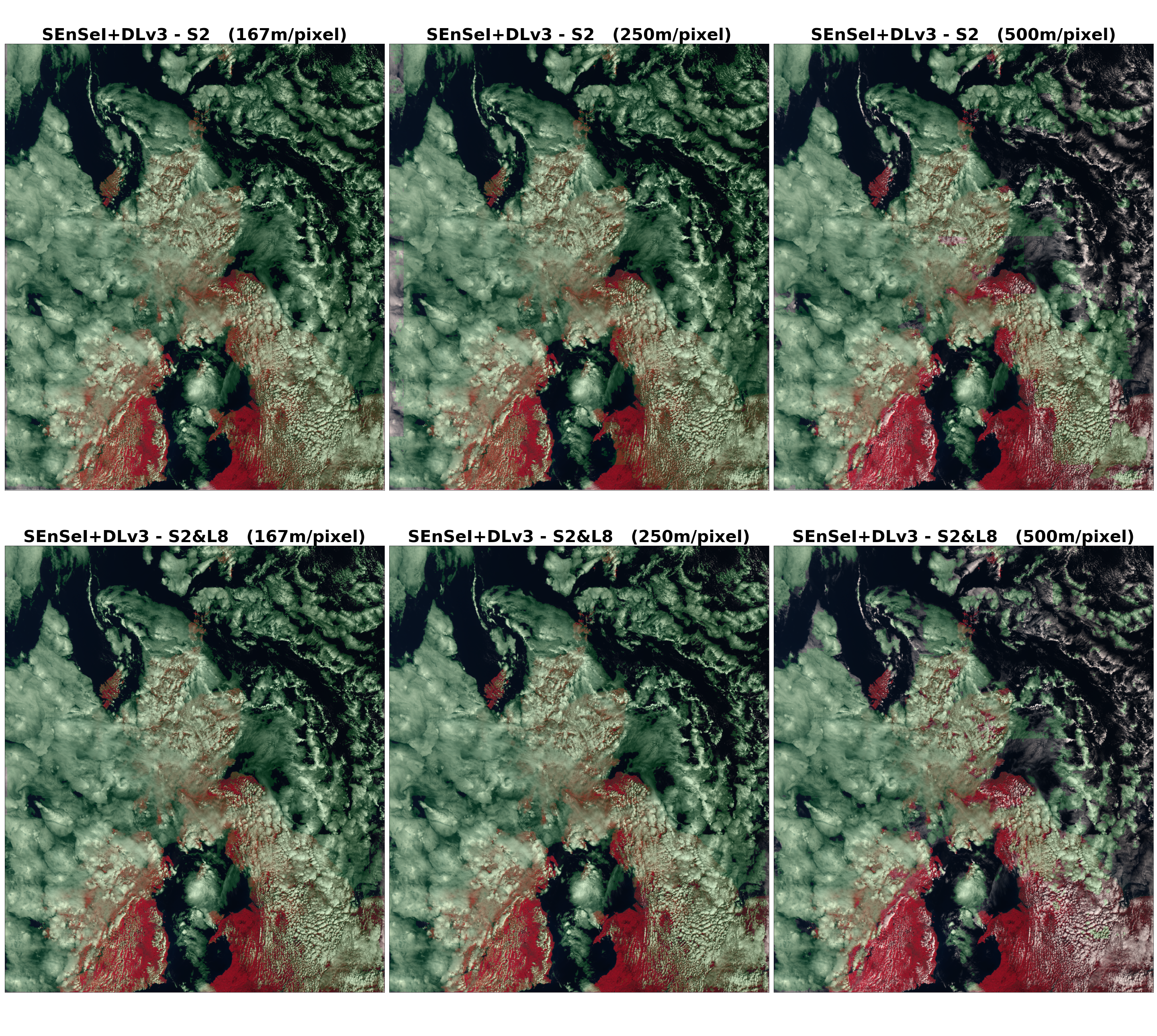}
\caption{Predictions of \textit{SEnSeI+DLv3 - S2} and \textit{SEnSeI+DLv3 - S2\&L8} for a Sentinel-3 scene over the UK and Ireland from August 2020. The scene has been cropped slightly for better visualisation. Here we display it in false colour with S3 as Red, S2 as Green, and S1 as Blue. The green overlay used for detections does not necessarily indicate correct detection, because we do not have a ground-truth for this scene. Masks produced at the original resolution of 500 m fail in areas with high frequency clouds, whereas at 250 m and 167 m, these are segmented more successfully. Some visible improvements in performance can be observed in the \textit{SEnSeI+DLv3 - S2\&L8} predictions, e.g. the false detections over Cardigan Bay, Wales made by the model trained only on Sentinel-2 are avoided. This figure is best viewed digitally in colour.}\label{fig:slstr-vis}
\end{figure*}

Finally, we present results over a scene from Sentinel-3's SLSTR. We use the 6 bands S1--S6, as these have close equivalents in Sentinel-2 and Landsat 8 from Green into the SWIR. They are all at 500 m resolution, and can be easily converted into reflectance values using \textit{SNAP}'s rad2refl preprocessing tool. We predicted clouds on the nadir-view stripe ``A'' data, and present the results in Fig. \ref{fig:slstr-vis}. As in our experiment using Per\'uSat-1, we predict clouds for different effective resolutions. This time, however, we upsample the images, artificially raising their resolutions from 500 m, to 250 m and 167 m. This was done because the original 500 m data contains very high-frequency clouds not seen in the 20--30 m data the model was trained on. What is more, because DeepLabv3's output is interpolated over an output stride of 8, it would not be able to segment these very small clouds successfully.

Whilst the results for SLSTR are only visual, we believe they show great promise. At higher spatial resolutions, especially, the models seem able to accurately segment clouds, at a resolution roughly 5-10 times coarser than they were trained on, up to 250 m/pixel. Whilst the development and the testing of SEnSeI focused on independence with respect to spectral band combinations, these results from SLSTR (and those from Per\'uSat-1) show SEnSeI-based models can work well across large resolution ranges, which massively increases the number of sensors they can be applied to without any retraining.

\section{Discussion}
\label{sec:discuss}

\subsection{Findings}
\label{subsec:findings}

In this work we have introduced the concept of sensor independence with respect to machine learning models in remote sensing, and applied it successfully to cloud masking, with both a neural network and a large convolutional model. The experimental results in Section \ref{sec:results} not only showcase state-of-the-art cloud masking performance in our models, but also the novel property of sensor independence. Sensor independence is an opportunity for deep learning's strength (assimilation of information from massive amounts of training data) to be combined with thresholding methods' strength (universality over many sensors, through their use of physical measurements).

In developing SEnSeI, we have produced a highly performant cloud masking algorithm that achieved Overall Accuracy greater than 92\% on 3 test sets from 3 different satellites, Sentinel-2, Landsat 8, and Per\'uSat-1, and with visually promising results on Sentinel-3's SLSTR instrument, all without any retraining whatsoever. We believe this proven compatibility with multiple sensors, beyond those used in training, is a novel result in the field of deep learning for remote sensing.

The experimental results do not unambiguously show SEnSeI to always offer an advantage to models in terms of statistical performance. Whilst the experiment on Landsat 8 demonstrated that sensor independence \textit{can} lead to improvements in performance, the Sentinel-2 experiment also showed that sometimes a transfer learning approach (as we define it in Section \ref{subsec:cloudmasking}) can be more successful. In both cases, though, the use of aggregated training data from multiple satellites was shown to be preferable to using data from a single satellite. Based on this, we argue that further work into sensor independence---and interoperability more generally---is vital for continuing to improve performance and expand usability of deep learning models in remote sensing.

Throughout Section \ref{sec:results} we have pointed to how the results correspond to our three hypotheses. Here, we consider each hypothesis in turn and pull together the outcomes of our experiments in relation to them:

\textit{\textbf{Hyp. A:} ``If SEnSeI is added to a deep learning model, there is no reduction in performance when trained and tested on data from a single sensor''.}

Whilst there is some good evidence for this hypothesis in our results, it is somewhat nuanced. The six DeepLabv3+ models tested on Sentinel-2---four with SEnSeI and two without---show a spread of results which fall slightly outside the $\pm0.5\%$ range that was measured using cross-validation on one of the models. Clearly, the increase in performance from \textit{DLv3 - S2} to \textit{SEnSeI+DLv3 - S2 [fixed bands]} shows that adding SEnSeI does not necessarily cause a drop in performance, with the OA actually increasing by 0.42\%, though this is just below what we deem to be a statistically significant difference of 0.5\%.

However, the drop in performance for \textit{SEnSeI+DLv3 - S2} is significant, at around 1\%---and is consistent with the 1\% drop seen for between \textit{NN - S2} and \textit{SEnSeI+NN - S2}. Consider, though, that this model has been trained to produce cloud masks on any combination of 3 or more Sentinel-2 bands, which is a much more general problem than one in which all bands are always present. So, we believe this experiment is evidence that SEnSeI has no deleterious effects on performance \textit{when the training task is the same}, but that there is a small cost to performance (1\%) when using SEnSeI to produce a more generalised, sensor independent model.

\textit{\textbf{Hyp. B:} ``If SEnSeI is added to a deep learning model, the model can be trained on multiple sensors without a drop in performance relative to when it is trained on an individual sensor.''}

To fully appreciate whether this hypothesis holds, we must consider two cases. First, what effect adding Landsat 8 data in training has on Sentinel-2 test performance, and second, the reciprocal case, when Sentinel-2 data is added, what affect is seen on Landsat 8 test performance. In the prior case, the performance of \textit{SEnSeI+DLv3 - S2\&L8} actually shows a moderate increase (0.65\% in OA, 0.63\% in BA, and 0.63\% in $F_1$ score) over the equivalent model trained on Sentinel-2 alone, \textit{SEnSeI+DLv3 - S2}. This gain in performance, though, does not quite make up for the initial decrease of 1\% seen when adding SEnSeI with random band combinations. 

In the second case, on Landsat 8 test data, the results are even more positive. The performance of \textit{SEnSeI+DLv3 - S2\&L8 (COMMON)} is significantly higher than the performance of the specialised, non-sensor independent Landsat 8 model, \textit{DLv3 - L8}. OA is 1.21\% higher, BA is 7.66\% higher, and $F_1$ 6.25\% higher. This seems to have been driven by the models trained only on the Landsat 8 CCA dataset (including those published by others) having highly imbalanced Precision and Recall values. This may be because SPARCS does not have the same sampling distribution as the Landsat 8 CCA dataset, as it focuses on clouds which are not extended across the whole scene, and which have complex and diverse structures. This difference in the data distribution  between training and test sets appears to lead to models which are too `picky' and miss many of the SPARCS dataset's clouds. The introduction of Sentinel-2 data at training has produced more balanced predictions on the SPARCS dataset, with Precision and Recall much more equal. This is promising evidence for the power of sensor independence; training on data from more sources, with different labellers and sensors, produces models which are more robust to changes in data sampling, and predictions which draw on information from a wider range of situations. Therefore, we believe the Landsat 8 SPARCS results strongly support the hypothesis, and indeed exceeds it, by showing in this instance that performance on one sensor, Landsat 8, can be improved by training data from another sensor, Sentinel-2.

\textit{\textbf{Hyp. C:} ``SEnSeI can enable a model to perform adequately on a previously unseen sensor, given training data that provide close equivalents to the unseen sensor's spectral bands.''}

Several times in this work, models are shown to perform well on sensors not included in training. First, let us consider the performance of \textit{SEnSeI+DLv3 - S2} in the experiment on the Landsat 8 SPARCS dataset. Across several band combinations, OA remained above 90\%, with the highest being for the band combination using all possible bands (except thermal bands, which were not available during training and so cannot be included in inference). This model achieved the highest BA of any model (93.05\%), and had very high recall but low precision---the inverse of the models trained solely on the Landsat 8 CCA dataset.

Landsat 7 presented difficulties for our models regarding Hyp. (C), which shows the often unpredictable nature of deep learning models when used in unfamiliar settings. In the case of Landsat 7, we believe that the significant saturation over high reflectance areas (often clouds) led to a flat texture that the model was unfamiliar with, and thus struggled to correctly classify. As we see in Fig. \ref{fig:L7-vis}, high reflectance, saturated areas (e.g. the thick clouds in the 2nd row in Fig. \ref{fig:L7-vis}) are sometimes mis-classified. Another issue is a difference in annotation style, with some thin clouds that the model detected not being included in the ground-truth (e.g. the thin clouds not annotated as such in the 4th row in Fig. \ref{fig:L7-vis}).

Meanwhile, at resolution scales outside those that the model was trained on, SEnSeI performed surprisingly well. For the finer resolution data in the CloudPeru2 dataset, sensor independent models achieved OA consistently above 92\% at 5.6 m, with a likely drop off in performance in the range of 3--5 m per pixel, with predictions at 2.8 m showing a roughly 1\% decrease in performance relative to 5.6 m results. 

At the coarser resolution range of Sentinel-3's SLSTR, no suitable labelled dataset could be found, however using our model on a product over the UK shows that visually impressive predictions can be made, although only by artificially upsampling the scene, which is somewhat inefficient. Predictions seem to become less reliable beyond 250 m/pixel, with the model failing to produce good results at 500 m/pixel. We think this limitation is related to the structure of clouds at this resolution, with the high frequency changes in cloud cover being too fine for DeepLabv3+'s output stride of 8 pixels to segment properly, leading to an averaging over class predictions and thus either all-cloud or all-clear predictions over areas with both clear and cloudy pixels.

\subsection{Dataset Design}

One factor which greatly affects the practicality of sensor independence, is the homogeneity of labelling styles, and data sampling schemes, between datasets. In some of our experiments, differences in annotation style may have caused drops in performance, notably for Landsat 7. This does not mean either style of annotation was ``wrong'' per se, but rather that they have chosen to define what is and isn't included in the category of `cloud' differently. In other experiments, such as those using Landsat 8 SPARCS as a test set, we believe the imbalanced Precision and Recall of some models compared to others is evidence that the data sampling scheme and class distribution of training data can have significant impacts on performance on a fixed test set. This issue was also noted by~\cite{Jeppesen2019rsnet}.

Where possible, we believe datasets should be designed with a focus on compatibility with prior efforts, both in terms of their labelling styles, and in their file formats and structure. Not only will this help to accelerate sensor independent model design by reducing the time spent on preprocessing work for each sensor and dataset, but will also allow studies with non-sensor independent designs to rapidly train and test on multiple datasets. Additionally, dataset creators should provide as much information as possible on the sampling of the data (how it was selected, what the edge cases are in class labels, etc.), so that users can understand what a model's performance in a given dataset is representative of.

\section{Conclusions \& Future Work}
\label{sec:conc}

We propose SEnSeI as a possible route to building sensor independence into existing models for remote sensing applications. By treating each spectral band as a member of a set of all possible bands, and using a permutation invariant network design, SEnSeI  translates any satellite's measured reflectances and brightness temperatures into a common representation. The framework is presented in the context of cloud masking, and is shown to work well with both a simple neural network, and the large convolutional model, DeepLabv3+. It enables DeepLabv3+ to simultaneously predict clouds over five different satellites without retraining, and using all available bands. Furthermore, the performance in Landsat 8 was significantly higher when trained on data from Sentinel-2 as well as Landsat 8, showing that a better model comes when using training data sampled from multiple sensors.

The sensor independent models appeared to struggle when given data that is very unlike that used in training (e.g. Landsat 7 with its saturation, or resolutions very different from those in training). Nevertheless, we demonstrated our model---not trained on any Landsat 7 data---had a comparable accuracy to purpose-built models~\cite{Scaramuzza2011CCA} (Table \ref{tab:exp-L7}). Future work could improve matters further by adding information beyond spectra to the inputted descriptors, e.g. saturation levels, Point Spread Function, noise, or resolution. This could allow models to better handle images with different statistics to those used in training, and help to expand training sets to include even more sensors.

Training a model to be sensor independent appears to cause some penalty to performance in some circumstances (e.g. in the Sentinel-2 experiment). However, this is not always the case, as the sensor independent model outperformed others on Landsat 8. The additional usability and versatility of the sensor independent model, being shown to work on many other satellites for which it was not directly trained, is the key advantage of sensor independence. Robustness to unfamiliar sensors is not only essential when training data does not exist for those satellites, but can also be incredibly helpful if deep learning is to be used operationally on-board future missions, where real training data cannot be created prior to launch, or in ground segment processing, where algorithms are usually developed before downlinking of data begins.

In this work, we have focused on the problem of cloud masking. Many of the datasets we used also contain cloud shadow annotations, presenting an avenue for future research. For many users of cloud masks, shadows are as important to detect as the clouds themselves. As well as cloud masking, cloud removal is also a topic of significant interest, e.g.~\cite{Meraner2020cloud,Sarukkai2020cloud}. With data fusion techniques, cloudy areas can be in-filled with synthetic multispectral imagery generated from other data sources which are unaffected by cloud cover. In the future, our work could be extended to this challenge.

More widely, we believe we have shown that sensor independence is a worthwhile strategy for remote sensing tasks, not limited to cloud masking. We expect the greatest benefit to come for sensors with little annotation data, which can now benefit from the extensive annotations on other sensors. We plan to apply SEnSeI to a greater variety of remote sensing computer vision tasks in the future, including, for example, object detection and classification. SEnSeI's architecture is not specific to the problem of cloud masking, and therefore it should be possible to use it without modification for other tasks and with other models.

Finally, we wish to emphasise that the main thrust of this paper is not the raw performance of our models. Whilst comparisons between competing methods are an important part of machine learning research, we believe our findings invite collaboration, not competition. SEnSeI is an accessory which can be added to existing models, making them sensor independent (although future work should seek to ascertain why it did not work well with CloudFCN, specifically). For many computer vision tasks in remote sensing, we see limited adoption of published techniques in real-world settings, even when performance is high~\cite{Thompson2021realizing}. This is often because the models are highly specialised to a specific sensor or dataset, and cannot be deployed beyond it. We suggest that pursuing sensor independence---with SEnSeI or something like it---can give models real-world utility, which may ultimately prove to be more important than further increasing their performance.

% or 


%\appendices

%\section{}

\appendices

\section{Computational Efficiency}\label{appendix:computation}

SEnSeI is a lightweight model with a fairly small footprint in terms of both memory and computation time. Table~\ref{tab:sensei-params} gives a breakdown of the tuneable parameters in SEnSeI's different modules. An experiment was also conducted to measure the computation time for the models used in this paper, and to measure how the addition of SEnSeI affected this computation. Each model was given a 257-by-257 pixel scene 1,600 times. For variants with SEnSeI, different numbers of bands were given, but for models without SEnSeI, 13 bands was used, as this is the number of bands which Sentinel-2 data contains. The number of parameters for each model, and the time taken per megapixel, is given in Table~\ref{tab:computation}. SEnSeI adds a small amount of computation time to the models, although when only using a few bands, SEnSeI has no noticeable affect.

\begin{table}[!h]

\caption{Number of parameters per module in SEnSeI. The names of the modules correspond to those in Fig.~\ref{fig:sensei}. The band multiplication block uses a fixed offset of 0.5, and so has no trainable parameters.} \label{tab:sensei-params}
\centering
\begin{tabular}{cc}
\textbf{Module}         & \textbf{No. Parameters} \\ \hline
DESCRIPTOR BLOCK & 24'208 \\
PERMUTATION BLOCK & 62'848 \\
COMBINED DESCRIPTOR BLOCK & 123'488 \\
BAND MULTIPLICATION BLOCK & 0 \\
\textbf{TOTAL} & \textbf{210'544}
\end{tabular}

\end{table}

\begin{table}[!h]

\caption{Measured computational efficiency across both the 2-layer neural network and DeepLabv3+. SEnSeI adds roughly 30 ms/Megapixel to either model, when 13 spectral bands are inputted.} \label{tab:computation}
\centering
\begin{tabular}{ccc}
\textbf{Model}         & \textbf{No. Parameters} & \textbf{Speed (ms/Megapixel)} \\ \hline
NN                     & 1154                    & 36.8                              \\
SEnSeI+NN (3 bands)    & 2.27e5                  & 33.2                              \\
SEnSeI+NN (8 bands)    & 2.27e5                  & 51.6                              \\
SEnSeI+NN (13 bands)   & 2.27e5                  & 71.4                              \\
DLv3                   & 2.11e6                  & 94.6                              \\
SEnSeI+DLv3 (3 bands)  & 2.34e6                  & 88.4                              \\
SEnSeI+DLv3 (8 bands)  & 2.34e6                  & 103                               \\
SEnSeI+DLv3 (13 bands) & 2.34e6                  & 121                              
\end{tabular}

\end{table}

\section{Detailed Results on Sentinel-2}
\label{appendix:results}
As well as measuring the metrics over the total Sentinel-2 test sets, we also subdivided the test set based on the categorical tags assigned to each image (see Table \ref{tab:appendix}). Relating the various models' performance with the content of the images gives a more nuanced view of how they can be expected to behave in the real world.

\begin{table*}

\renewcommand{\arraystretch}{1.9}
\setlength{\tabcolsep}{3.5pt}

\caption{DeepLabv3+ model results over the Sentinel-2 test set (257 scenes). We also sub-divide the total test by the presence of several categories of surface type and cloud properties, described in Section \ref{subsubsec:S2dataset}. Categories with fewer than 30 subscenes are omitted. The best result for each metric and image category is boldened.}
\label{tab:appendix}
\centering

\begin{tabular}{cccccccccccccccccc}
\begin{tabular}{c} \textbf{\large{OA}} \end{tabular} &    \rot{\textbf{Total test set}} &  \rot{\textbf{forest/jungle}} &  \rot{\textbf{snow/ice}} &  \rot{\textbf{agricultural}} &  \rot{\textbf{desert/barren}} &  \rot{\textbf{shrublands}} &  \rot{\textbf{open\_water}} &    \rot{\textbf{low}} &   \rot{\textbf{high}} &   \rot{\textbf{thin}} &  \rot{\textbf{thick}} &  \rot{\textbf{isolated}} &  \rot{\textbf{extended}} &  \rot{\textbf{cumulus}} &  \rot{\textbf{stratocumulus}} &  \rot{\textbf{cirrus}} &  \rot{\textbf{ice\_clouds}} \\ \hline
DLv3 - S2 &  94.04 & 92.52 & 86.05 & 92.22 & 94.88 & 95.31 & 94.14 &  \textbf{94.92} &  92.97 &  91.10 &  95.26 & 89.54 & 93.92 & \textbf{95.00} & 96.42 & 91.93 & 95.30 \\
\makecell{DLv3 - S2\&L8 \\ {[}common bands{]}} & \textbf{95.19} & \textbf{93.28} & \textbf{89.71} & \textbf{95.07} & 95.85 & \textbf{95.92} & 93.83 & 94.82 & \textbf{94.52} & \textbf{92.28} & \textbf{95.93} & \textbf{90.48} & \textbf{94.83} & 94.93 & 96.63 & \textbf{93.84} & 96.85 \\
 \makecell{SEnSeI+DLv3 - S2 \\ {[}fixed bands{]}}   &  94.46 & 91.60 & 88.36 & 92.61 & \textbf{95.94} &              94.56 & 94.01 &  94.15 & 93.43 &  91.11 &  95.27 &     88.95 &     94.64 &    94.20 & 95.93 & 92.86 & 96.58 \\ 
 \makecell{SEnSeI+DLv3 - S2}   &  93.08 &          90.92 &     87.14 &         91.35 &          93.16 &              93.81 &       93.61 &  93.69 &  92.07 &  90.43 &  94.31 &     89.15 &     92.99 &    93.86 &                      95.78 &   91.33 &       93.66 \\
 \makecell{SEnSeI+DLv3 - S2\&L8}   &  93.73 &          90.70 &     84.87 &         90.20 &          93.77 &              93.78 &       \textbf{94.31} &  94.66 &  92.49 &  89.97 &  95.68 &     88.90 &     94.23 &    94.74 &                     \textbf{96.86} &   91.79 &       \textbf{97.45} \\
 %\makecell{Standard L1C \\ Product Mask}  &  83.11 &          75.83 &     73.81 &         82.09 &          83.59 &              83.33 &       76.50 &  76.80 &  87.63 &  77.31 &  82.07 &     71.37 &     82.74 &    78.06 &                      79.05 &   88.19 &       92.27 \\ 

\Xhline{3\arrayrulewidth}

\begin{tabular}{c} \textbf{\large{BA}} \end{tabular} &&&&&&&&&&&&&&&&& \\ \hline
                         DLv3 - S2 &  94.03 & 91.35 & 85.94 & 91.60 & 96.43 & 94.31 & 94.12 &  94.56 &  91.05 &  91.16 &  94.05 &     88.58 &     89.99 &    \textbf{94.57} &                      94.08 &   89.42 &       74.35 \\ 
\makecell{DLv3 - S2\&L8 \\ {[}common bands{]}} & \textbf{95.18} & \textbf{91.94} & \textbf{89.41} & \textbf{93.61} & \textbf{97.21} & \textbf{94.75} & 93.84 & 94.09 & \textbf{93.33} &\textbf{ 92.37} & 93.82 & \textbf{89.60} & \textbf{89.27} & 94.12 & 93.67 & \textbf{92.46} & 73.15 \\
 \makecell{SEnSeI+DLv3 - S2 \\ {[}fixed bands{]}}   &  94.42 &          90.44 &     88.49 &         91.04 &          96.38 &              92.69 &       94.01 &  93.54 &  90.81 &  91.03 &  93.16 &     88.25 &     90.21 &    93.48 &                      92.35 &   89.72 &       78.89 \\
 \makecell{SEnSeI+DLv3 - S2}   &  93.10 &          89.10 &     87.20 &         89.21 &          93.07 &              90.57 &       93.60 &  93.37 &  91.00 &  90.67 &  93.10 &     87.89 &     89.93 &    93.46 &                      93.05 &   89.99 &       80.03 \\
 \makecell{SEnSeI+DLv3 - S2\&L8 }   &  93.73 &          88.43 &     85.04 &         87.75 &          92.31 &              89.87 &       \textbf{94.27} &  \textbf{94.57} &  91.30 &  90.26 &  \textbf{94.64} &     87.23 &     91.27 &    94.51 &                      \textbf{94.46} &   90.06 &       \textbf{82.88} \\
 %\makecell{Standard L1C \\ Product Mask}   &  83.29 &          70.62 &     76.09 &         76.03 &          80.31 &              74.59 &       76.15 &  79.40 &  84.64 &  78.03 &  84.42 &     65.00 &     75.31 &    80.45 &                      81.28 &   83.56 &       71.50 \\ 

 \Xhline{3\arrayrulewidth}

\begin{tabular}{c} \textbf{\large{Precision}} \end{tabular} &&&&&&&&&&&&&&&&& \\ \hline
                         DLv3 - S2 &  94.57 &          91.64 &     83.03 &         85.03 &          62.24 &              90.33 &       94.67 &  96.79 &  96.71 &  93.81 &  97.69 &     85.09 &     98.80 &    96.96 &                      98.45 &   96.48 &       98.63 \\
\makecell{DLv3 - S2\&L8 \\ {[}common bands{]}} & \textbf{95.61} &\textbf{93.76} & \textbf{88.83} & \textbf{93.44} & 67.07 & 92.39 & 93.25 & 96.09 & \textbf{97.66} & \textbf{94.84} & 97.22  & \textbf{86.43} & 98.61 & 96.32 & 98.25 & \textbf{97.66} & 98.54 \\
 \makecell{SEnSeI+DLv3 - S2 \\ {[}fixed bands{]}}   &  94.58 &          90.00 & 84.55 & 88.23 &          \textbf{68.00} &              90.61 &       93.78 &  95.95 &  96.37 &  93.20 &  97.02 &     83.33 &     98.79 &    96.10 &                      97.88 &   96.36 &       98.87 \\
 \makecell{SEnSeI+DLv3 - S2}   &  94.00 &          91.36 &     83.47 &         86.98 &          55.35 &              92.28 &       94.14 &  96.24 &  97.03 &  94.09 &  97.40 &     85.37 &     98.87 &    96.45 &                      98.19 &   97.01 &       98.99 \\ 
 \makecell{SEnSeI+DLv3 - S2\&L8}   &  94.42 &          92.89 &     80.26 &         85.27 &          58.12 &              \textbf{94.23} &       \textbf{95.97} &  \textbf{97.15} &  97.07 &  93.95 &  \textbf{97.94} &     86.25 &     \textbf{99.00} &    \textbf{97.16} &                      \textbf{98.51} &   96.90 &       \textbf{99.08} \\ 
% \makecell{Standard L1C \\ Product Mask}   &  86.88 &          73.14 &     63.46 &         75.07 &          30.61 &              74.88 &       91.76 &  91.88 &  94.35 &  85.48 &  96.17 &     63.94 &     96.98 &    92.51 &                      96.79 &   94.23 &       98.51 \\ 
 
 \Xhline{3\arrayrulewidth}

\begin{tabular}{c} \textbf{\large{Recall}} \end{tabular} &&&&&&&&&&&&&&&&& \\ \hline
DLv3 - S2 &  94.17 & \textbf{87.18} &     85.18 & \textbf{90.03} & 98.30 & 92.18 & 93.30 &  95.60 &  94.34 &  90.81 &  96.20 & 85.31 & 94.59 & 95.72 & 97.35 & 93.45 & 96.51 \\
\makecell{DLv3 - S2\&L8 \\ {[}common bands{]}} & \textbf{95.31} & 87.15 & 87.19 & 89.95 & \textbf{98.83} & \textbf{92.24} & \textbf{94.24} & \textbf{96.20} & 9\textbf{5.38} & \textbf{91.82} & \textbf{97.57} & \textbf{86.62} & \textbf{95.77} & \textbf{96.30} & 97.81 & 94.67 & 98.22 \\
 \makecell{SEnSeI+DLv3 - S2 \\ {[}fixed bands{]}}   &  94.99 &          86.27 &     \textbf{89.46} &         87.08 &          96.91 &              88.70 &       94.02 &  95.32 &  95.29 &  91.51 &  96.92 &     85.86 &     95.40 &    95.43 &                      97.36 &   \textbf{94.77} &       97.61 \\
 \makecell{SEnSeI+DLv3 - S2}   &  92.89 &          82.60 &     87.65 &         83.84 &          92.95 &              83.67 &       92.75 &  94.30 &  92.83 &  89.28 &  95.26 &     83.60 &     93.50 &    94.54 &                      96.87 &   92.14 &       94.45 \\
 \makecell{SEnSeI+DLv3 - S2\&L8}   &  93.73 &          80.32 &     86.27 &         81.58 &          90.56 &              81.54 &       92.28 &  94.84 &  93.34 &  88.59 &  96.49 &     81.56 &     94.74 &    95.13 &                      \textbf{97.82} &   92.84 &       \textbf{98.29} \\
% \makecell{Standard L1C \\ Product Mask}   &  80.27 &          51.98 &     93.08 &         60.75 &          76.39 &              55.96 &       57.25 &  71.83 &  89.75 &  73.81 &  80.23 &     43.33 &     84.00 &    74.03 &                      78.15 &   91.01 &       93.47 \\ 
 
 \Xhline{3\arrayrulewidth}
 
\begin{tabular}{c} \large{$\mathbf{F_1}$} \end{tabular} &&&&&&&&&&&&&&&&& \\ \hline
                         DLv3 - S2 & 94.37 &  89.35 & 84.09 & 87.46 & 76.22 & 91.25 & 93.98 & \textbf{96.20} & 95.51 & 92.28 & 96.94 & 85.20 & 96.65 & \textbf{96.33} & 97.90 & 94.94 & 97.56 \\
\makecell{DLv3 - S2\&L8 \\ {[}common bands{]}} & \textbf{95.46} & \textbf{90.33} & \textbf{88.00} & \textbf{91.66} & 79.91 & \textbf{92.32} & 93.74 & 96.15 & \textbf{96.50} & \textbf{93.30} & \textbf{97.40} & \textbf{86.53} & \textbf{97.17} & 96.31 & 98.03 & \textbf{96.14} & 98.38 \\
 \makecell{SEnSeI+DLv3 - S2 \\ {[}fixed bands{]}}   & 94.79 & 88.10 & 86.94 & 87.65 & \textbf{79.92} & 89.64 &       93.90 &  95.64 &  95.83 & 92.35 &  96.97 &     84.58 & 97.06 &    95.76 &  97.62 &  95.56 & 98.23 \\
 \makecell{SEnSeI+DLv3 - S2}   &  93.44 &          86.76 &     85.51 &         85.38 &          69.38 &              87.76 &       93.44 &  95.26 &  94.88 &  91.62 &  96.32 &     84.47 &     96.11 &    95.49 &                      97.52 &   94.52 &       96.66 \\
 \makecell{SEnSeI+DLv3 - S2\&L8}   &  94.07 &          86.15 &     83.16 &         83.38 &          70.80 &              87.43 &       \textbf{94.09} &  95.98 &  95.17 &  91.19 & 97.21 &     83.84 &     96.82 &    96.13 &                      \textbf{98.16} &   94.83 &       \textbf{98.68} \\
% \makecell{Standard L1C \\ Product Mask}   &  83.45 &          60.77 &     75.47 &         67.16 &          43.71 &              64.05 &       70.51 &  80.62 &  92.00 &  79.22 &  87.48 &     51.65 &     90.02 &    82.25 &                      86.48 &   92.59 &       95.92 \\ 
 
 \Xhline{3\arrayrulewidth} 
 
No. of scenes & 257 & 42 & 41 & 41 & 34 & 40 & 66 & 129 & 126 & 110 & 156 & 90 & 137 & 140 & 92 & 104 & 51 \\
Cloud \%	& 53.05 & 36.02 & 43.29 & 30.14 & 8.34 & 26.53 & 49.06 & 67.21 & 79.22 & 58.60 & 78.09 & 35.30 & 92.74 & 68.64 & 85.74 & 81.09 & 97.27 \\
 
\end{tabular}
\end{table*}

Of the metrics used, OA and Recall are relatively unaffected by the ratio of pixels which are cloudy, whereas BA, Precision and $F_1$ are all highly dependent on the class distribution. Therefore, when inspecting these results, it is also important to consider the percentage of cloudy pixels that are present in a given category of images, and only compare results across categories in metrics where it is reasonable to do so.

\section*{Acknowledgments}

The authors would like to thank the many researchers whose datasets and models were used in this work, including those at the USGS for the Landsat datasets, and Giorgio Morales and CONIDA for the CloudPeru2 dataset. We would also like to thank Pierre-Philippe Mathieu and colleagues at the $\Phi$-lab, ESRIN for giving continued support throughout this project. This project was funded through STFC Grant 1912521, and ESA Contract 4000130413/20/I-DT.

\ifCLASSOPTIONcaptionsoff
  \newpage
\fi




\bibliographystyle{IEEEtran}
\bibliography{IEEEabrv,./main}
%\begin{thebibliography}{1}


%\end{thebibliography}


\begin{IEEEbiography}[{\includegraphics[width=1in,height=1.25in,clip,keepaspectratio]{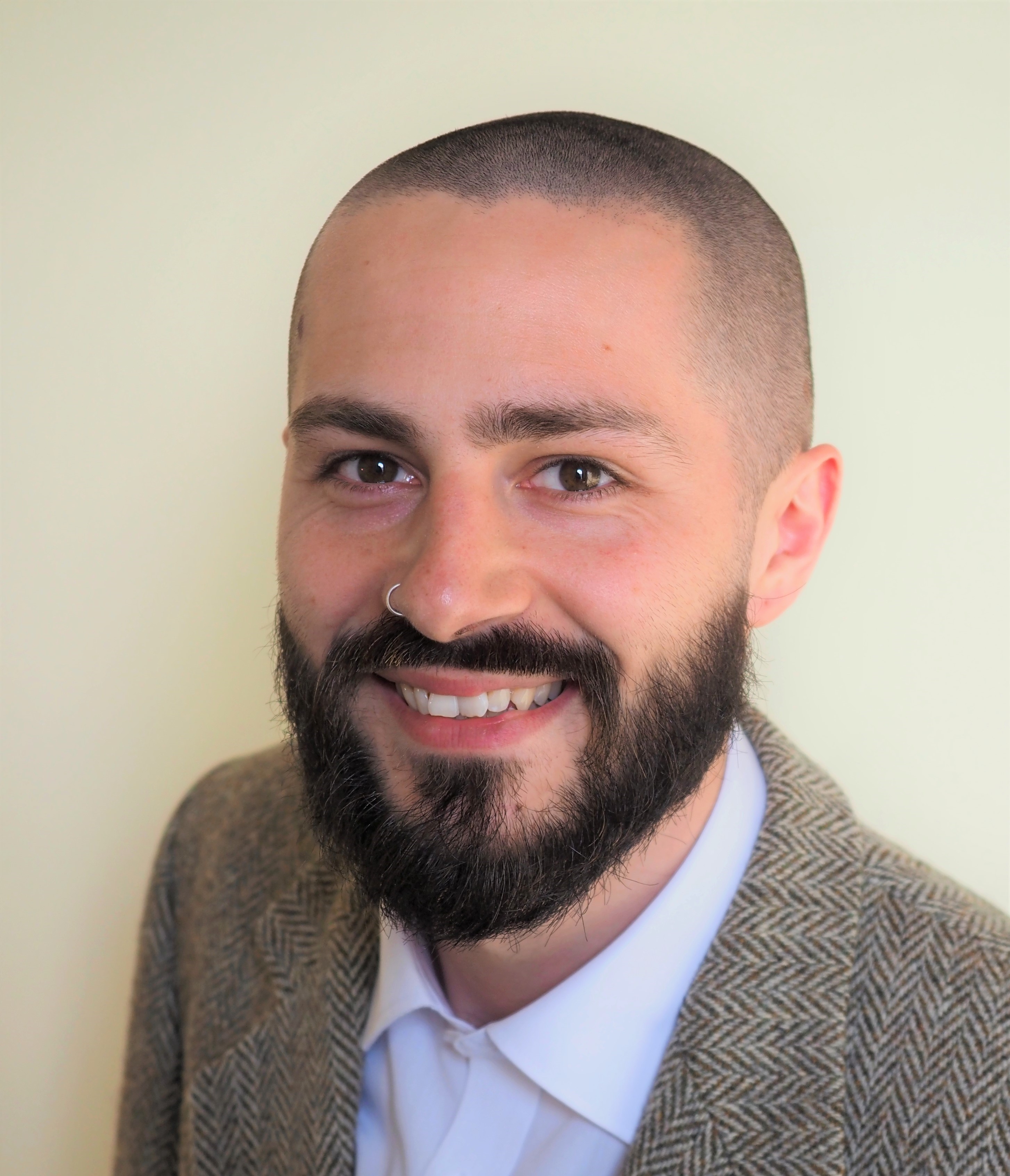}}]{Alistair Francis}
was born in Maryport, UK, in 1994. He received a B.A. Hons in Physics from the University of Oxford in 2016. He was then awarded an M.Sc. in Space Science \& Engineering from University College London in 2017, and completed his Ph.D. at Mullard Space Science Laboratory, University College London, in 2021.

He has previously worked on projects with Createc Ltd., Cockermouth, UK, from 2015--2016, and Cortexica Ltd., London, UK, in 2018. Since October 2021, he holds a postdoctoral research fellowship at ESRIN's $\Phi$-lab, Frascati, Italy.

His research interests include machine learning and computer vision applications in remote sensing data of Earth and Mars, semi-automated data labelling, and problems involving data fusion across multispectral, hyperspectral, and SAR instruments.

\end{IEEEbiography}

\begin{IEEEbiography}[{\includegraphics[width=1in,height=1.25in,clip,keepaspectratio]{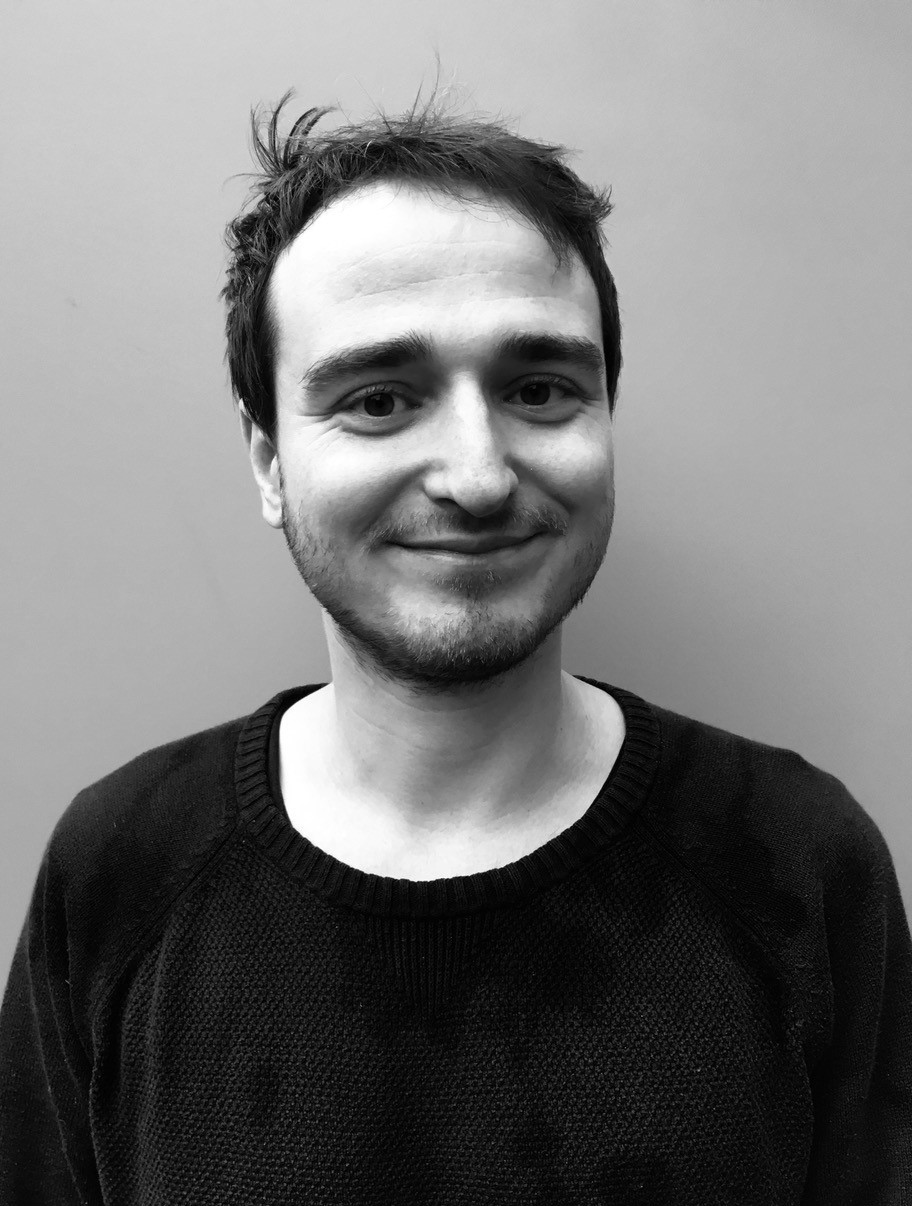}}]{John Mrziglod}
received the B.Sc. and M.Sc. degrees in Meteorology from the Universit\"at Hamburg, Germany, in 2014 and 2018,  respectively. 

He spent one and a half years at the $\Phi$-Lab of the European Space Agency, Rome, Italy, as Young Graduate Trainee for Digital  Earth Observation Technologies, after which he was an Earth Observation Data Analyst at the UN World Food Programme, Rome, Italy. He is currently a Video Game Developer in Hamburg, Germany.

His research interests include the application of deep learning to remote sensing and climate data.
\end{IEEEbiography}

%\newpage

\begin{IEEEbiography}[{\includegraphics[width=1in,height=1.25in,clip,keepaspectratio]{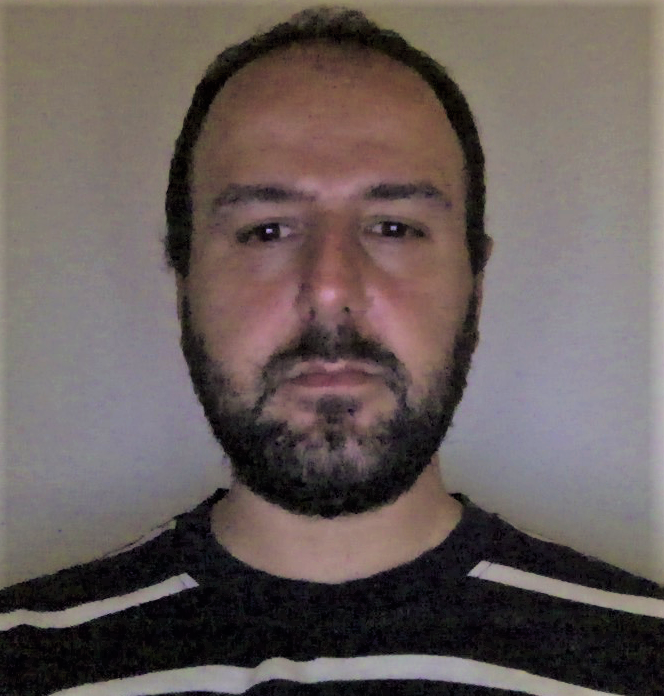}}]{Panagiotis Sidiropoulos}
received his Ph.D. degree in artificial intelligence from the Centre for Vision, Speech, and Signal Processing, University of Surrey, Surrey, UK, in 2012. 

He is currently an honorary lecturer in the Mullard Space Science Laboratory, UCL, while holding a Head of Data and Knowledge Science position in industry. His work is in the intersection of academic excellence and commercial applicability, having co-authored more than 100 papers while receiving $>$£5 million in commercial funding. 

His research interests include the use of machine learning, computer vision and data science techniques in remote sensing applications.
\end{IEEEbiography}

\begin{IEEEbiography}[{\includegraphics[width=1in,height=1.25in,clip,keepaspectratio]{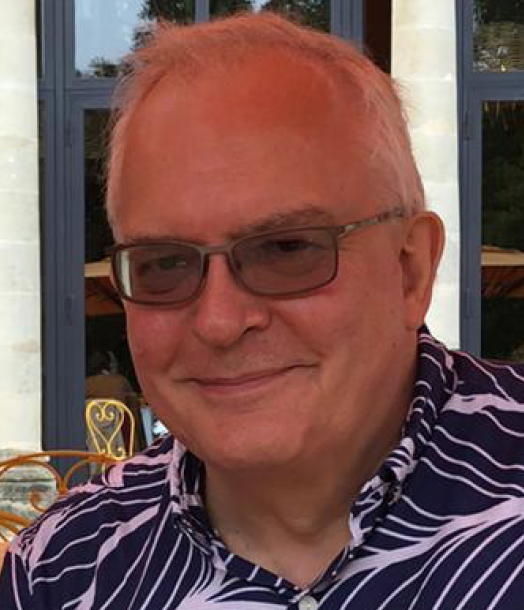}}]{Jan-Peter Muller}
received a B.Sc. degree (Hons.) in Physics from the University of Sheffield, U.K., in 1976; the M.Sc. and D.I.C. degrees in atmospheric physics and dynamics from Imperial College London, U.K., in 1977, and a Ph.D. degree in planetary meteorology from the University of London, UK, in 1982.

He is currently Professor of image understanding and remote sensing with University College London (UCL), where he has been on the faculty since 1984. He is head of the Imaging Group at the Mullard Space Science Laboratory, which is part of the Department ofSpace and Climate Physics at UCL. He was a member of the MODIS science team from 1989 until 2016 and is an active member of the MISR science Team since 1990. He is a member of the ESA Harmony mission MAG responsible for developing automated methods for the retrieval of 3D cloud-top heights, winds and cloud presence.

His research interests include the practical applications of machine vision and deep learning to remote sensing of the earth and planets, with a special focus on 3-D imaging and super-resolution restoration technologies for applications in climate modelling, weather forecasting, and planetary exploration.
\end{IEEEbiography}


%\vfill


\end{document}